%% file: main.tex
\documentclass[10pt,journal,compsoc]{IEEEtran}
\usepackage{amsmath}
\usepackage[utf8]{inputenc}
\usepackage{amssymb}
\usepackage{adjustbox}
\usepackage{xcolor}
\usepackage{tikz}

\usepackage{graphicx}
\usepackage{subcaption}
\usepackage{supertabular}
\usetikzlibrary{calc}

\usepackage{todonotes}
\usepackage[backend=biber]{biblatex}
\presetkeys{todonotes}{color=blue!20}{}

\begin{document}

\title{Event Camera Guided Visual Media Restoration \& 3D Reconstruction: A Survey}

\author{Aupendu~Kar, Vishnu~Raj, and Guan-Ming~Su%
    \IEEEcompsocitemizethanks{
        \IEEEcompsocthanksitem All authors are with Dolby Laboratories, Inc.
        \IEEEcompsocthanksitem Aupendu Kar (email: \texttt{aupendu.kar@dolby.com})\\
        Vishnu Raj (email: \texttt{vishnu.raj@dolby.com})\\
        Guan-Ming Su (email: \texttt{guanmingsu@ieee.org})
    }
}

\IEEEtitleabstractindextext{%
\begin{abstract}
Event camera sensors are bio-inspired sensors which asynchronously capture per-pixel brightness changes and output a stream of events encoding the polarity, location and time of these changes. These systems are witnessing rapid advancements as an emerging field, driven by their low latency, reduced power consumption, and ultra-high capture rates. This survey explores the evolution of fusing event-stream captured with traditional frame-based capture, highlighting how this synergy significantly benefits various video restoration and 3D reconstruction tasks. 
The paper systematically reviews major deep learning contributions to image/video enhancement and restoration, focusing on two dimensions: temporal enhancement (such as frame interpolation and motion deblurring) and spatial enhancement (including super-resolution, low-light and HDR enhancement, and artifact reduction). This paper also explores how the 3D reconstruction domain evolves with the advancement of event driven fusion. Diverse topics are covered, with in-depth discussions on recent works for improving visual quality under challenging conditions. 
Additionally, the survey compiles a comprehensive list of openly available datasets, enabling reproducible research and benchmarking. 
By consolidating recent progress and insights, this survey aims to inspire further research into leveraging event camera systems, especially in combination with deep learning, for advanced visual media restoration and enhancement.
\end{abstract}

}

\maketitle
\IEEEdisplaynontitleabstractindextext
\IEEEpeerreviewmaketitle

\input{10_intro}

\input{20_working}

\input{30_temporal_domain}

\input{40_spatial_domain}

\input{50_3dreconstruction}

\input{80_datasets}
\input{90_conclusion}

\printbibliography

\end{document}

%% file: 10_intro.tex
\section{Introduction} \label{sec:intro}

Vision is the dominant human sense for perceiving the world and, in combination with the brain, facilitates learning. Traditionally, frame-based RGB cameras have been the preferred choice for visual sensing, as they capture rich color, texture, and semantic information. However, they struggle in challenging conditions such as extreme lighting and fast relative motion. Event cameras, also known as Dynamic Vision Sensors~\cite{brandli2014240,gallego2020event}, operating on principles similar to the human retina, offer a novel approach to capturing changes in a visual scene. Inspired by biological vision, event cameras capture asynchronous pixel-wise intensity changes, offering advantages such as low latency, high speed, and high dynamic range.

Event camera systems offer unique advantages over traditional frame-based cameras through their novel approach to capturing visual information~\cite{gallego2020event}.
Unlike conventional frame-based cameras that record absolute intensity at fixed intervals, event cameras asynchronously detect per-pixel logarithmic brightness changes ($\Delta L/\Delta t$), generating sparse spatiotemporal event streams with microsecond temporal resolution ($>10,000$ fps) and high dynamic range ($>120 dB$).
These attributes uniquely position them to address critical limitations in traditional imaging systems—motion blur in high-speed scenarios, over/under exposure in extreme lighting, and bandwidth inefficiencies in redundant static scenes.
By capturing the changes in the scene asynchronously, they allow for high temporal resolution and reduced motion blur, making them particularly suitable for image and video restoration applications under extreme capture conditions.

Since the event camera sensor only captures the changes in brightness asynchronously, it concurrently loses the grayscale information of the scene. Traditionally, the grayscale information is captured by an active pixel sensor, which is co-located with the event sensor in the same chip~\cite{brandli2014240}. Downstream tasks utilize event integration, i.e., the fusion of the event data stream with grayscale capture for image reconstruction. Bao et al.~\cite{bao2024temporal} introduce a novel method for intensity image creation using the information about the time of event emission for each pixel. Lei et al.~\cite{lei2024many} study the impact of the number of events needed for image reconstruction using event cameras. They study fixed duration and fixed number of events as their primary reconstruction strategies and provide insights into the trade-off between temporal resolution and image quality for each strategy.

\begin{figure*}[!t]
    \centering
    \includegraphics[width=0.90\textwidth]{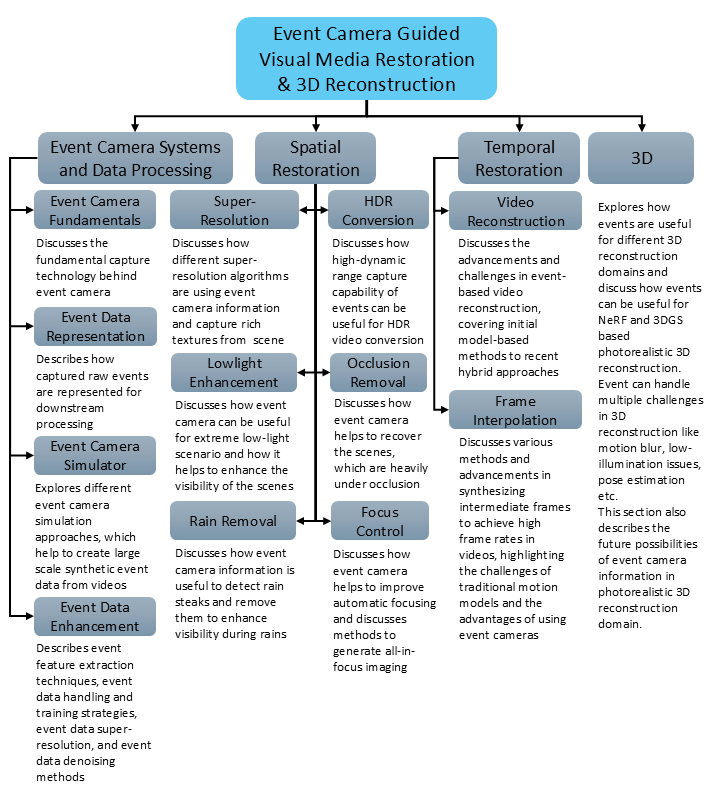}
    \caption{Organization of the study.}
    \label{fig:paper_map}
\end{figure*}

Recent advancements in deep learning have significantly accelerated event camera applications by developing architectures tailored to their asynchronous, sparse data streams. Transformers~\cite{vaswani2017attention} and spiking neural networks (SNNs)~\cite{tavanaei2019deep} can process event-based spatiotemporal features directly, enabling tasks like \emph{high-speed video reconstruction}~\cite{liu2017high} and \emph{low-light enhancement}~\cite{liu2023low}. 
For instance, hybrid SNN-CNN pipelines~\cite{lee2020spike} use spiking layers to extract microsecond-level temporal cues, fused with CNNs for spatial super-resolution, achieving better optical flow estimation than frame-based methods.
Physics-informed networks further integrate event generation models (e.g., $ \Delta L = \nabla L \cdot v + \partial L/\partial t $) as differentiable layers, improving optical flow estimation accuracy in extreme lighting. 
Attention mechanisms, such as cross-modal asymmetric transformers, dynamically weight event and RGB features to reduce motion blur artifacts in deblurring tasks~\cite{sun2022event}. 
These methods leverage event cameras’ $\mu$-second-resolution temporal data to overcome traditional vision bottlenecks, making them highly beneficial for applications in autonomous navigation, computational photography, and real-time augmented reality. Deep learning can unlock unprecedented visual fidelity in dynamic environments by aligning architectural innovations with event cameras’ innate strengths - low latency, high dynamic range, and motion robustness.

\textbf{Motivation.} In this survey paper, we focus on the recent developments in using event cameras for image and video restoration applications. There exist excellent survey papers~\cite{gallego2020event,chakravarthi2024recent,zheng2023deep,shariff2024event} covering the working principles and broad applications of event camera systems. Compared to them, in this survey, we intend to focus on the restoration and enhancement capabilities offered by event camera sensors when used in conjunction with traditional frame-based cameras. We explore different image and video capture challenges, such as low-light enhancement, HDR reconstruction, artifact reduction, and focus control, as well as how it unlocks new capabilities, such as 3D reconstruction.

\textbf{Outline.} We start the survey by presenting a primer on Event camera systems. After discussing the working principles, we discuss event data simulators and event data enhancement methods to round up the data acquisition task. In the following two sections, we present how event cameras can be leveraged to enhance visual content in spatial and temporal domains. Next, we discuss how event camera data can be leveraged for 3D scene reconstruction and survey recent work leveraging event data for the same. Finally, we include a comprehensive list of publicly available event camera datasets with the hope that curious readers may find it interesting to experiment with event data streams. We provide a map of this survey in Figure~\ref{fig:paper_map}.

%% file: 20_working.tex
\section{Event Camera Systems and Data Processing} \label{sec:working}

\input{20_1_working}

\input{20_2_event_rep}

\input{20_3_eventenhance}

%% file: 20_1_working.tex
\subsection{Event Camera Fundamentals} 
\label{subsec:fundamentals}

Event cameras are bio-inspired vision sensors that capture visual information as asynchronous events triggered by changes in brightness. Different from traditional cameras, which capture images at fixed sampling intervals (e.g., $30$ fps), event camera systems output a continuous stream of events, each containing a pixel address, timestamp, and polarity (increase or decrease) of the brightness change.

Each pixel in an event camera operates independently, continuously comparing its current brightness level to a stored reference level. If the difference exceeds a predefined threshold, the pixel generates an event and resets its reference level to the current brightness. This process allows event cameras to capture motion and changes in the scene with very low latency (on the order of microseconds) and a high dynamic range (typically $120$ dB).

The working principle of an event camera at a circuit level can be summarized as follows:
\begin{enumerate}
    \item Incoming light is processed by each pixel independently, continuously, and asynchronously.
    \item The photo-diodes convert the incoming light into an electrical current, which is subsequently transformed into a corresponding voltage signal.
    \item This voltage signal is compared to a reference voltage on a logarithmic scale to detect any changes in light intensity.
\end{enumerate}

\begin{figure}[htbp]
    \centering
    \begin{subfigure}{0.45\textwidth}
        \centering
        \includegraphics[width=\linewidth]{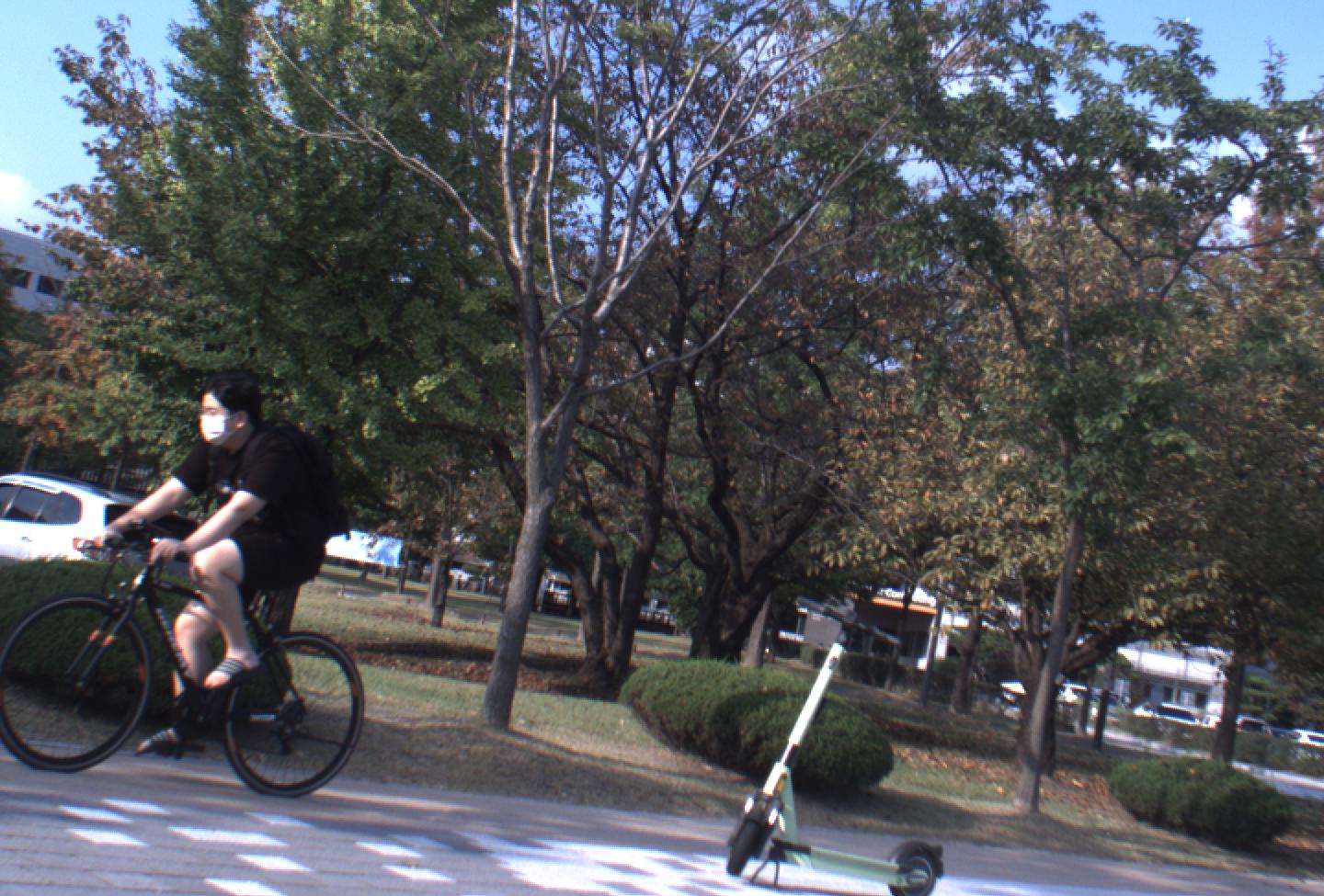}
        \caption{Captured intensity frame}
        \label{fig:ex_rgb}
    \end{subfigure}
    \hfill
    \begin{subfigure}{0.45\textwidth}
        \centering
        \includegraphics[width=\linewidth]{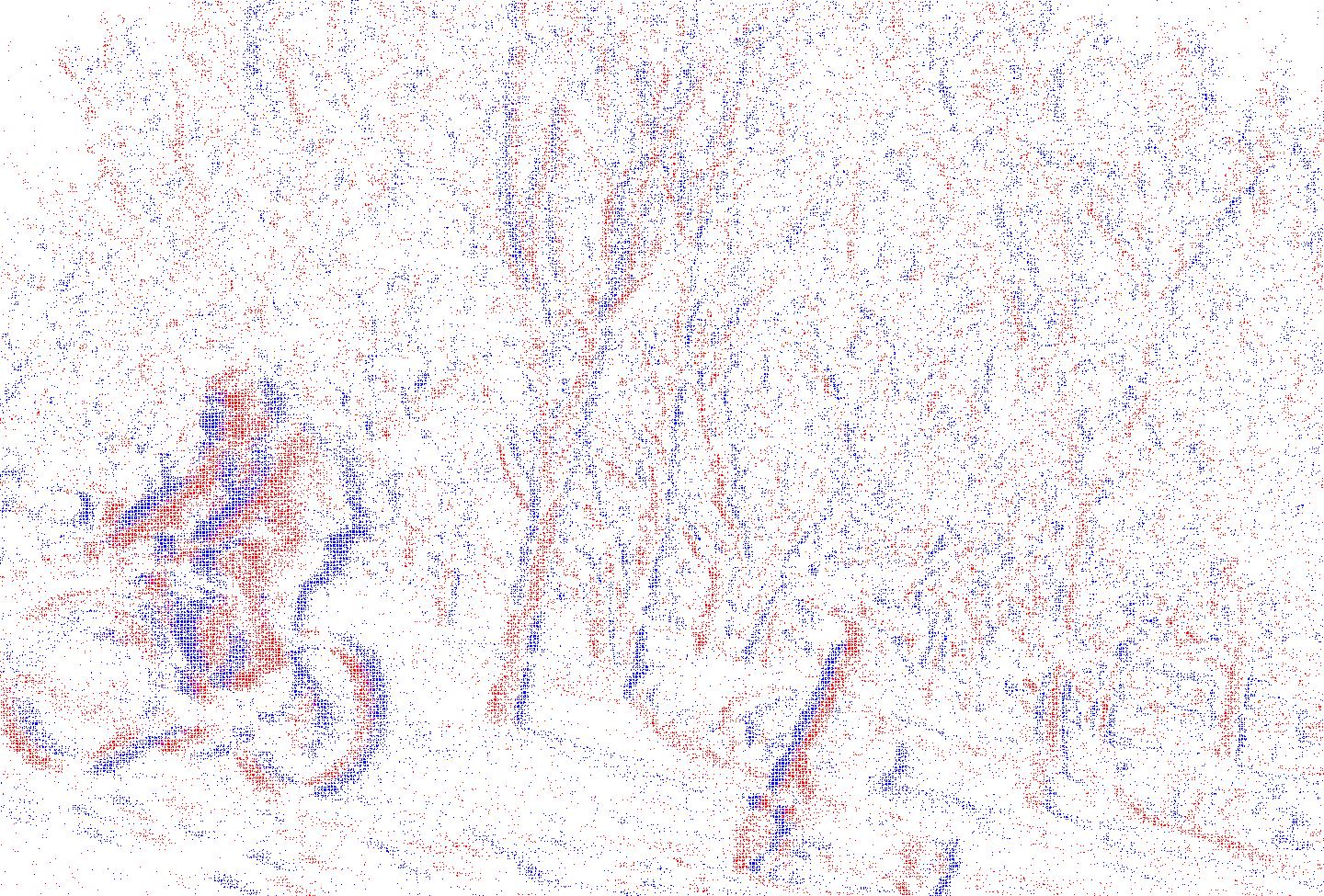}
        \caption{Captured event frame}
        \label{fig:ex_event}
    \end{subfigure}
    \caption{An example of a captured intensity frame and its corresponding events between this captured intensity frame and the next frame. Red and blue dots in the event frame represent negative and positive polarity, respectively. The continuous event data stream is projected into a 2D frame for visualization.}
    \label{fig:example_images}
\end{figure}

Formally, each event can be interpreted as a tuple $(x, y, p, t)$, where $(x, y)$ represents the pixel location, $t$ the timestamp, and $p \in \{-1, +1\}$ is the polarity indicating the direction of brightness change. An event is triggered whenever a change in the logarithmic intensity $L$ surpasses a predefined threshold $C$. This can be represented as
\begin{align}
    p = \begin{cases}
        +1, &\qquad L(x, y, t) - L(x, y, t - \Delta t) \geq C \\
        -1, &\qquad L(x, y, t) - L(x, y, t - \Delta t) \leq -C \\
        0, &\qquad \text{otherwise}
    \end{cases}
\end{align}
where $\Delta t$ is the time interval since the last event at pixel $(x, y)$. We can represent the change in brightness since the last event at a pixel by
$
    \Delta L (x, y, t) = L(x, y, t) - L(x, y, t - \Delta t),
$
and hence, the $\Delta L(x, y, t) = p \cdot C$. Further, a stream of events can be represented as
\begin{align}
    \mathcal{E} = \{ e_i \}_{i=1}^{N} = \{ (x_i, y_i, p_i, t_i) \} \qquad i \in N.
    \label{eqn:event-stream}
\end{align}

An example image of event camera capture is given in Figure~\ref{fig:example_images}. In Figure~\ref{fig:ex_rgb}, a frame captured by an RGB camera is shown where the camera is moving from left to right and the cyclist is moving from right to left. Figure~\ref{fig:ex_event} shows the events captured over a small interval of time from the same scene. In the event data stream, red represents the regions where intensity went down during capture, and blue represents the regions where intensity went up. As we can see, in the parked e-scooter, as the white frame of the e-scooter moves from right to left, blue dots show up on the front side of the e-scooter, followed by red dots on the backside of the main axle, indicating how the high-intensity white region moved across in the frame. Similarly, as the black t-shirt cyclist moves fast from left to right, we see trailing blue dots signifying an increase in intensity due to the dark color of the t-shirt covering the regions, and leading blue dots signifying bright skin covering the regions. For the purpose of visualization, we have projected the event data stream across time into a 2D plane, where the changes are visible.
We provide a detailed comparison of a traditional RGB camera with an event camera in Table~\ref{tab:comp-rgb-event}. 

\begin{table*}[t]
\begin{adjustbox}{width=\textwidth}
\centering
\begin{tabular}{lll}
    \multicolumn{1}{c}{\textbf{Characteristics}} & 
    \multicolumn{1}{c}{\textbf{RGB Camera}} & 
    \multicolumn{1}{c}{\textbf{Event Camera}} \\
    \hline
    Capture method          & Fixed rate full frame & Asynchronous per-pixel \\
    Temporal resolution     & milliseconds & microseconds \\
    Latency                 & High, due to full frame capture & low, due to asynchronous events  \\
    Data output             & Fixed-frame rate with absolute brightness values  & Sparse data stream encoding brightness changes \\
    Dynamic range           & Moderate, around $60dB$  & Very high, around $140dB$ \\
    Power consumption       & High, due to continuous processing & Low, due to sparse processing \\
    Applications            & General purpose imaging, video & High-speed vision, HDR imaging
\end{tabular}
\end{adjustbox}
\caption{Comparison of traditional RGB camera with Event camera.}
\label{tab:comp-rgb-event}
\end{table*}

Event camera systems focus on changes in light intensity rather than absolute light levels, and hence, only relevant information regarding a scene is captured, thus minimizing redundancy. Further, since the event camera detects changes in logarithmic scale rather than absolute values, common issues like overexposure and underexposure, prevalent in traditional systems, can be avoided.

%% file: 20_2_event_rep.tex
\subsection{Event Representation}   
\label{subsec:event_representations}

The sparse nature of the event data stream (Eqn.~\ref{eqn:event-stream}) makes it difficult to apply the DNN models predominantly designed for frame-based cameras. This necessitates the development of alternative representation formats for event data that can capture the visual information power appropriately. Some of the popular event representation methods~\cite{zheng2023deep} are described below.

\noindent
\textbf{Image Based Representation.} 
Stacking asynchronous events into a synchronous 2D image representation, similar to frame-based cameras, is a straightforward solution to adapt events to existing DL methods. It should be noted that these representations are often set to preserve polarities, timestamps, and event counts. Popular stacking approaches are
\begin{enumerate}
    \item Based on polarity, two separate channels can be used to evaluate the histogram of positive and negative events. The output of the event camera is collected into frames over a specified time interval $T$, using a separate channel for the event polarity. This can be finally fused into synchronous time events~\cite{maqueda2018event}.
    \item Based on timestamps, the events can be aggregated into synchronous frames to capture holistic information~\cite{wang2019event, deng2020amae, bai2022accurate}. The time duration of the event stream is divided into $n$ equal-scale portions, and the $n$ grayscale frames are formed by merging the events in each interval through pointwise summation. These grayscale frames are further stacked together and are presented as input to the model.
    \item Based on the number of events, ~\cite{hu2020learning, messikommer2020event} propose a strategy to sample and stack events in a fixed constant number. By only updating the locations where new events are recorded, this method, when used in conjunction with sparse convolutions, can also maintain the temporal sparsity of events.
\end{enumerate}

\noindent
\textbf{Surface Based Representation.} 
Surface representation aims to map the event streams to a time-dependent surface and tracks the activity around the spatial location of the latest event. Usually, time $t$ is represented as a monotonically increasing function of position $(x, y)$. Surface of active events~\cite{benosman2013event} captures a time surface that encodes the time context in a neighbourhood region of the event and helps to maintain spatio-temporal information for downstream tasks. Because of the monotonically increasing nature of timestamps, different normalization methods are developed for temporal invariant data representations~\cite{lagorce2016hots, alzugaray2018ace, afshar2019investigation}.

\noindent
\textbf{Voxel Based Representation.} 
Voxel representation maps the raw events into the nearest temporal grid within temporal bins. By partitioning both space and time into discrete levels, voxel-based representation quantizes and accumulates the events into voxel grids in $(x, y, t)$ space. A pioneering work in this direction, Zihao et al.~\cite{zihao2018unsupervised} propose to insert events into volumes using a linearly weighted accumulation to improve the resolution along the temporal domain. To compactly capture raw information temporal spikes with minimal information loss, Baldwin et al.~\cite{baldwin2022time} propose the concept of time-ordered recent event volume. The graph is constructed by voxelizing the spatio-temporal event, subsampling voxels to get representative voxels as vertices, and finally integrating internal events in a voxel along the time axis to obtain the node features. The proposed EV-VGCNN is capable of finding neighbors and calculating edge weights for vertices. Choudhary et al.~\cite{choudhury2025triplane} proposed a novel triplane and probabilistic autoencoder framework for compact and unified event stream representation, featuring a two-stage training scheme and Poisson-based voxel encoding that enables efficient reconstruction and compatibility with diffusion models.

\noindent
\textbf{Graph Based Representation.} 
Graph representation transforms the sparse events within a time window into a set of connected nodes. For the problem of object detection, Bi et al.~\cite{bi2020graph} propose a residual graph CNN architecture to obtain a compact graph representation. Each node is a sampled event from a time-space voxel grid, and nodes are connected with edges only if they have a weighted spatio-temporal Euclidean distance. Deng et al.~\cite{deng2022voxel} propose a lightweight voxel graph CNN, which has been shown to achieve high accuracy with low model complexity.

\noindent
\textbf{Spike Based Representation.} 
Spike representation uses SNNs to extract features from event streams asynchronously to solve diverse tasks~\cite{gu2020tactilesgnet, orchard2015hfirst}. However, complex dynamics and the non-differentiable nature of spiking neural networks severely limit the applicability of these methods to large-scale problems.

\noindent
\textbf{Learning Based Representation.} 
Learning based representation leverages the power of data to learn optimal representations to convert an asynchronous event stream to flexible representations that can be used for diverse tasks. Popular methods utilize Multi-Layer Perceptron (MLP) architectures to aggregate spatio-temporal information~\cite{gehrig2019end}, Long-Short Term Memory (LSTM)~\cite{cannici2020differentiable} to integrate information into the temporal axis, etc. Because it is fully differentiable, this method allows for the extraction of the most relevant representation for downstream tasks from the data itself. Huang et al.~\cite{huang2025neural} present a learning-based framework for efficiently representing event voxel grids, which are typically sparse and inefficient to store due to the spatiotemporal nature of event camera data. Three neural representations - MLP, Tensor Decomposition, and Hash Encoding - along with their respective strengths and limitations in enabling deep learning models to process event data for vision tasks. Choudhary et. al \cite{choudhury2025modeling} introduce a modified neural field model, composed solely of a multi-layer perceptron (MLP), to effectively represent sequences of event frames generated by event cameras, where each pixel encodes the polarity of brightness change within a temporal window. The more advanced learning based event feature extraction techniques and architectures are discussed in Section~\ref{subsubsec:data_proc}

\subsection{Event Camera Simulators} 
\label{subsec:simulators}

While event camera systems offer unparalleled advantages in terms of high dynamic range, no motion blur, and asynchronous sensing, their scarcity and high non-production-scale hardware cost remain challenging for the research community when acquiring event camera data streams. Hence, obtaining synthetic data for exploratory research and algorithm validation in a controlled and cost-efficient setting is necessary.
This has led to the development of several event camera simulator systems that can generate large amounts of affordable and reliable event data.

Based on the underlying principle of event simulation, modern event simulators are broadly classified into three major groups:

\noindent
\textbf{Optimization Based Simulators.}
Optimization-based simulators operate based on optimizing hand-crafted modules for generating event data~\cite{rebecq2019events, palinauskas2023generating, ziegler2023real, hu2021v2e}. ESIM~\cite{rebecq2019events} can generate events, standard images, and inertial measurements while simulating arbitrary camera motion in 3D scenes. It also provides complete ground truth data, including camera pose, velocity, depth maps, and optical flow maps. However, ESIM does not perform noise modeling and is limited to synthetic 3D scenes or high-framerate input video. V2E~\cite{hu2021v2e} generates realistic synthetic events from intensity frames and addresses non-idealities such as Gaussian event threshold mismatch and intensity-dependent noise using handcrafted modules. Palinauskas et al.~\cite{palinauskas2023generating} introduce an event simulator for a robotics use case by extending ESIM~\cite{rebecq2019events} into MuJoCo~\cite{todorov2012mujoco} platform-based simulation. By using different levels of frame interpolation method, Ziegler et al.~\cite{ziegler2023real} introduce an event camera simulator that can perform near-real-time simulation of event camera data.

\noindent
\textbf{Physical Based Simulators.}
Han et al.~\cite{han2024physical} and Lin et al.~\cite{lin2022dvs} try to incorporate physical law-based imperfections in the event sensors in addition to the changes in the intensity to simulate close-to-real event data streams. By taking into account the fundamental circuit properties, Lin et al.~\cite{lin2022dvs} develop a realistic event camera simulator that can faithfully model the voltage variations, randomness in photon receptions, and noise caused by leakage current into a stochastic process and showed that the simulated events strongly resemble real event data. By directly interfacing with a 3D scene, Han et al.~\cite{han2024physical} design a realistic lens simulation block and a novel multi-spectral rendering block to combine both optical as well as circuit-level imperfections. The results show that the system is able to produce high-fidelity data under different lighting conditions and motion speeds.

\noindent
\textbf{Learning Based Simulators.} 
Gu et al.~\cite{gu2021learn} and Zhang et al.~\cite{zhang2024v2ce} aim to leverage developments in deep learning to approximate the event generation process and hence make it a completely data-driven approach. This removes the elaborate manual efforts required for heuristically tuning each of the components in a traditional simulator, as well as offers more adaptiveness to more domains. Gu et al.~\cite{gu2021learn} introduce a method to learn pixel-wise distributions of event contrast thresholds for a given domain, enabling stochastic sampling and parallel rendering to generate event representations that closely match real event camera data. This is achieved through a novel divide-and-conquer discrimination scheme that adaptively assesses synthetic-to-real consistency based on local image and event statistics. This method has also been shown to have better domain adaptation capabilities. V2CE~\cite{zhang2024v2ce} introduces a two-stage method for realistic event data simulation using a 3D UNet backbone for predicting event voxels, followed by an event sampling module. The whole system is trained using specialized loss functions to enhance the quality of generated event voxels and is shown to be able to convert video to high-fidelity event streams with precise events.

%% file: 20_3_eventenhance.tex
\subsection{Event Data Enhancement} 
\label{subsec:event_enhancement}

Most of the video restoration and 3D reconstruction models use the event data to fuse it with the RGB sensor data to improve the performance. 
The characteristics of raw event data are different from conventional RGB camera captures. In order to achieve better visual media enhancements and reconstruction, the event data needs to be processed for better fusion with image modality. 
Therefore, researchers developed multiple methods to process the event data better. Another challenge in event cameras is noise and low-resolution capture. Generally, the event data is prone to noise, especially in low-light conditions, and therefore, denoising is required. The resolution of event cameras is low as compared to conventional RGB cameras. In order to fuse the low-resolution event data with high-resolution RGB camera data, super-resolution event data is necessary. In this section, we will discuss multiple strategies for efficient event data processing, event denoising, and event super-resolution in three different sections.

\subsubsection{Efficient Event Feature Extraction} 
\label{subsubsec:data_proc}

The event camera captures visual information through asynchronous events triggered by brightness changes. The captured event camera data is different from other imaging data. Therefore, special techniques for event cameras are required to process the event data. 
Huang et al.~\cite{huang2023eventpoint} propose a self-supervised method for detecting and describing local features in event streams and introduce a novel event stream representation method called Tencode. Tencode processes event data to obtain pixel-level interest points and descriptors through a neural network.
To achieve efficient event data processing, Sun et al.~\cite{sun2022menet} introduce the MENet model that features a dual-branch structure. Prior methods often ignore the motion continuity between adjacent windows, which leads to the loss of dynamic information and the extraction of more redundant information. MENet addresses these issues by enhancing feature extraction and reducing redundancy, making it a more efficient solution for event stream processing. This dual-branch includes a base branch for full-sized event point-wise processing and an incremental branch that captures temporal dynamics between adjacent spatiotemporal windows. The incremental branch, which is equipped with a point-wise memory bank, significantly reduces computational complexity and improves processing speed. Traditional event-based backbones often rely on image-based designs, which overlook the unique properties of event data, such as time and polarity. To address this, Peng et al.~\cite{peng2023get} introduce Group Event Transformer (GET), a novel vision transformer backbone specifically designed for event-based vision tasks, which decouples temporal-polarity information from spatial information throughout the feature extraction process to utilize the temporal and polarity information of events fully. It introduces a new event representation named Group Token, which groups asynchronous events based on their timestamps and polarities. 
The advantage of this new representation is that it decouples temporal-polarity information from spatial information throughout the feature extraction process. 
By decoupling temporal and polarity information at the token level, GET enables the Transformer’s attention mechanisms to operate more effectively across both spatial and temporal-polarity domains. This structured representation allows the Event Dual Self-Attention block to selectively attend to meaningful patterns.
To address the challenge of poor generalizability in existing deep neural networks when deployed at higher inference frequencies of events, Zubic et al.~\cite{zubic2024state} introduce a novel approach to event-based vision using State-Space Models (SSMs) with learnable timescale parameters. 
Transformers typically rely on fixed temporal windows and dense input representations. They struggle with generalization when deployed at different frequencies than those used during training. Transformers are less effective at capturing the continuous temporal dynamics inherent in event streams, leading to performance drops when the input frequency changes. 
Here, frequency refers specifically to the event representation sampling rate. This is the rate at which the asynchronous event stream is aggregated into a representation for processing.
For example, a 20 Hz frequency means events are grouped into 50 ms windows.
Higher frequencies (e.g., 100 Hz or 200 Hz) mean smaller time windows (10 ms or 5 ms), which allow for finer temporal resolution but also pose challenges like aliasing.
SSMs are designed to handle temporal sequences by maintaining a hidden state that evolves over time. This hidden state captures the underlying dynamics of the event stream, which allows the model to adapt to varying frequencies without needing retraining. The learnable timescale parameters in SSMs help to generalize better across different temporal resolutions and address the poor generalization issue seen in other models.
Unlike traditional methods that convert event data into dense image-like representations, FARSE-CNN~\cite{santambrogio2024farse} maintains the inherent sparsity and asynchronous nature of event data for efficient asynchronous event processing. The architecture combines recurrent and convolutional neural networks along with compression modules to learn hierarchical features in both space and time. It captures the dynamics of event streams while maintaining sparsity. 
FARSE-CNN can be considered an event representation technique as it processes and represents event data in a way that preserves its spatio-temporal sparsity.

\subsubsection{Event Data Training} 
\label{subsubsec:data_train}

Multiple training strategies have been developed to process event information and provide efficient training. Gallego et al.~\cite{gallego2019focus} introduce a collection and taxonomy of $22$ objective functions, termed Focus Loss Functions, to analyze events' alignment with intensity information in motion compensation approaches. The applicability of these loss functions is demonstrated across multiple tasks, including rotational motion, depth, and optical flow estimation, showcasing the potential of event cameras in various challenging scenarios. 
Sparse events are a challenge, and they often result in incomplete data, loss of crucial information, and difficulties in making accurate predictions. Sparse event completion is essential to enhance data quality by filling in the gaps, leading to more comprehensive and reliable datasets. This improves the accuracy of analysis and predictions, enabling better decision-making. Zhang et al.~\cite{zhang2024event} address the challenge of sparse event data.
The proposed method treats event streams as 3D event clouds in the spatiotemporal domain and employs a diffusion-based generative model to generate dense event clouds in a coarse-to-fine manner.
Yang et al.~\cite{yang2023event} propose a self-supervised learning framework through contrastive learning~\cite{chen2020simple, he2020momentum, khosla2020supervised, gao2021simcse} to pretrain a network using paired event camera data and natural RGB images for handling event camera data. 
Contrastive learning is an unsupervised machine learning technique where models learn to differentiate between similar and dissimilar pairs of data by maximizing the similarity of positive pairs and minimizing the similarity of negative pairs to understand and organize data without labeled examples. Yang et al.~\cite{yang2023event} use the pre-trained network in multiple event-driven downstream tasks and show promising performance.

There is a challenge of annotated data scarcity in event-based vision due to the recency of event cameras. To overcome this, Jian et al.~\cite{jian2023unsupervised} propose an unsupervised domain adaptation algorithm to reduce the distributional domain gap between frame-based annotated datasets and event-based data. The domain gap arises because event-based data and frame-based data have different characteristics and distributions. The paper aims to bridge this gap by using unsupervised domain adaptation techniques, allowing knowledge from frame-based annotated datasets to be transferred to event-based data for tasks such as image classification. Their technique tries to map the data from both domains into a shared domain-agnostic embedding space. This algorithm leverages contrastive learning and uncorrelated conditioning to transfer knowledge from conventional camera-annotated data to event-based data.
The labeled event data is costly and labor-intensive to annotate. Therefore, Klenk et al.~\cite{klenk2024masked} introduce Masked Event Modeling, a self-supervised learning framework, to reduce the dependency on labeled event data. The proposed method pre-trains a neural network on unlabeled events from any event camera recording. This pretraining significantly improves the accuracy of downstream tasks when the model is fine-tuned. 
Gehrig et al.~\cite{gehrig2022high} explore the necessity of high-resolution sensors in event cameras. The study reveals that, contrary to popular belief, low-resolution event cameras can outperform high-resolution ones in specific conditions, such as low illumination and high-speed scenarios. This is due to the higher per-pixel event rates in high-resolution cameras, which lead to increased temporal noise under these conditions. Lu et al.~\cite{lu2025rgb} introduce the first event-RAW paired and pixel-level aligned dataset for event-based image signal processing (ISP). This dataset includes $3373$ RAW images with a resolution of $2248 \times 3264$ and their corresponding events, spanning $24$ scenes with three exposure modes and three lenses. 
The approach involves using the event data to guide the image signal processing (ISP) pipeline, which includes steps like demosaicing, white balancing, denoising, and color correction. By integrating event data with RAW images, the ISP process can leverage the high temporal resolution and dynamic range of event data to enhance the quality of the resulting RGB images.

Yao et al.~\cite{yao2024exploring} investigate the susceptibility of Spiking Neural Networks (SNNs) to adversarial attacks, specifically targeting raw event data from Dynamic Vision Sensors (DVSs). These sensors capture visual information through asynchronous spikes triggered by brightness changes, offering high temporal resolution and low power consumption. The study introduces a novel adversarial attack approach that directly targets raw event data, addressing the challenges of three-valued optimization and the need to preserve data sparsity. The proposed method treats discrete event values as probabilistic samples, focuses on specific event positions to enhance attack precision, and employs a sparsity norm to maintain the original data's sparsity.

\subsubsection{Event Spatial and Temporal Up-Conversion}  
\label{subsubsec:super_res}

In low-brightness or slow-moving scenes, events are often sparse and noisy, which poses challenges for event-based tasks. To solve these challenges, Xiang et al.~\cite{xiang2022temporal} propose an event temporal up-sampling algorithm that generates more effective and reliable events by estimating the event motion trajectory using a contrast maximization algorithm and then up-sampling the events through temporal point processes.

The spatial low-resolution events directly impact the performance of event-based tasks like reconstruction, detection, and recognition. The spatial super-resolution of events leads to enhanced performance in downstream applications such as event-based visual object tracking and object detection, etc. It also helps to improve the quality of the reconstructed images. 
Li et al.~\cite{li2021event} propose a real-time framework based on a spiking neural network (SNN) to generate high-resolution event streams from low-resolution inputs and deploy it on a mobile platform. 
Spiking neural networks (SNNs) for event data are useful due to their alignment with the asynchronous, sparse nature of event streams. SNNs process data using discrete spikes, making them ideal for capturing fine-grained temporal dynamics. Their energy efficiency and ability to model spatiotemporal patterns enable effective super-resolution of low-resolution event streams.
Shariff et al.~\cite{shariff2024event} integrate binary spikes with Sigma Delta Neural Networks (SDNNs), leveraging a spatiotemporal constraint learning mechanism designed to simultaneously learn the spatial and temporal distributions of the event stream. Zhang et al.~\cite{zhang2024neuromorphic} introduce a self-supervised super-resolution prototype that adapts to any low-resolution input source without requiring prior training or side knowledge. This method leverages asynchronous streaming events to estimate high-resolution counterparts, significantly improving visual richness and clarity. Huang et al.~\cite{huang2024bilateral} propose a bilateral event mining and complementary network (BMCNet) that leverages a two-stream network to process positive and negative events individually. This approach allows for comprehensive mining of each event type and facilitates the exchange of information between the two streams through a bilateral information exchange (BIE) module. 
Zhang et al.~\cite{zhang2024crosszoom} introduce a unified neural network (CZ-Net) designed to address motion blur using low-resolution events in neuromorphic cameras. The network is a multi-scale blur-event fusion architecture that leverages the scale-variant properties of events and images to achieve cross-enhancement. This architecture effectively fuses cross-modal information, utilizing attention-based adaptive enhancement and cross-interaction prediction modules to mitigate distortions in low-resolution events.
Unlike traditional methods for event stream super-resolution, which often require high-quality, high-resolution frames, 
Weng et al.~\cite{weng2022boosting} propose a recurrent neural network (RNN) based method that does not rely on image frames and achieves super-resolution for a large-scale factor ($\times16$). 
Traditional methods often mix positive and negative events directly, leading to a loss of detail and inefficiency. Therefore, Liang et al.~\cite{liang2024efficient} introduce a Recursive Multi-Branch Information Fusion Network (RMFNet) to enhance the spatial resolution of event streams. The method separates positive and negative events to extract complementary information, followed by mutual supplementation and refinement.

\subsubsection{Event Denoising} 
\label{subsubsec:event_denoising}

Event cameras, which capture dynamic scenes with high temporal precision, offer advantages like high dynamic range, low latency, and low power consumption. However, they are prone to noise due to their differential signal output and logarithmic conversion. They often suffer from noise, particularly under low illumination and varying camera settings. The outputs of event cameras are prone to various types of noises such as photon shot noise, dark current shot noise, leakage current noise, and hot pixel noise~\cite{jiang2024edformer}. Therefore, event denoising becomes crucial for better visual media restoration and 3D reconstruction. In this direction, 
EventZoom~\cite{duan2021eventzoom} addresses the challenge of joint denoising and super-resolving neuromorphic events by leveraging a 3D U-Net backbone architecture, which is trained in a noise-to-noise fashion~\cite{lehtinen2018noise2noise} to enforce noise-free event restoration. 
In the noise-to-noise approach, both input and output are unfiltered noisy data of the same signal, and the target is to recover the signal from those noisy measurements.
If the noise is zero-mean and independent across samples, then the expected value of a noisy image is the clean image.
So, if we train a model to minimize the difference between two noisy versions of the same image, the model learns to predict the underlying clean signal.
Duan et al.~\cite{duan2023neurozoom} also employ a 3D U-Net backbone neural architecture to train in a noise-to-noise fashion~\cite{lehtinen2018noise2noise}.  
Shi et al.~\cite{shi2024polarity} introduce a novel dual-stage denoising method for event cameras. The proposed method, called Polarity-Focused Denoising (PFD), leverages the consistency of polarity and its changes within local pixel areas to handle noise effectively. Due to camera motion or dynamic scene changes, the polarity and its variations in triggered events are closely linked to these movements, which help to achieve effective noise management.
Zhang et al.~\cite{zhang2024neuromorphic} present an iterative coarse-to-fine approach, where an event-regularized prior provides high-frequency structures and dynamic features for blind deblurring, while image gradients help regulate noise removal. 

Jiang et al.~\cite{jiang2024edformer} propose the EDformer model that leverages the transformer architecture to learn spatiotemporal correlations among events, enabling effective denoising across varied noise levels. They introduce the ED24 dataset, which encompasses $21$ noise levels and noise annotations, providing a robust foundation for evaluating denoising algorithms. The data corresponding to $21$ different noise levels are defined based on $21$ different illumination conditions.
Duan et al.~\cite{duan2023neurozoom} also introduce a display-camera system for recording high frame-rate videos at multiple resolutions for data collection, and they trained a 3D-Unet for joint super-resolution and denoising. Display-camera system refers to a setup where a display screen shows high-frame-rate videos, and an event camera is positioned to record the display.
Duan et al.~\cite{duan2024led} also present a comprehensive dataset, the LED dataset, designed to address the challenges of denoising event camera data in real-world scenarios. They also propose a novel denoising framework, DED, that uses homogeneous dual events to better separate noise from signal events. 
Dual events refer to a dual-sampling setup, where two identical event cameras are used to simultaneously capture the same scene under identical conditions. This setup is designed to exploit the inconsistency of noise events and the consistency of signal events across two recordings.
Existing datasets and evaluation metrics for denoising are limited in scale and noise diversity, often relying on sensor information or manual annotation. To address these limitations, Ding et al.~\cite{ding2023mlb} present the E-MLB dataset, which includes $100$ scenes with $4$ noise levels, making it $12$ times larger than the largest existing dataset. They also introduce a comprehensive benchmark designed to evaluate denoising algorithms for event-based cameras and propose a novel non-reference denoising metric called the Event Structural Ratio (ESR), which measures the structural intensity of events independently of the number of events and projection direction. 

\subsubsection{Multi-Modal and Security Processing} 
\label{subsubsec:other_event}

Lin et al.~\cite{lin2023e2pnet} introduce E2PNet, the first learning-based method for registering event data to 3D point clouds. A 3D point cloud is a collection of data points in a three-dimensional coordinate system, representing the external surface of objects or scenes, typically constructed from a 3D scene using techniques like LiDAR or depth sensors. Registering event data to a point cloud is essential for enhancing the accuracy and robustness of scene understanding. Event data, which is more resilient to changes in lighting and motion, when aligned with the spatial information in the point cloud, provides a comprehensive analysis of dynamic scenes. This integration facilitates better object detection, tracking, and recognition, making the combined data more reliable and useful for various applications. The core of E2PNet is the Event-Points-to-Tensor (EP2T) network, which encodes event data into a 2D grid-shaped feature tensor. This representation allows the use of established RGB-based frameworks for event-to-point cloud registration without altering hyperparameters or training procedures.
Lin et al.~\cite{lin2022autofocus} address the challenge of focus control in event cameras. Traditional autofocus methods are ineffective for event cameras due to differences in sensing modality, noise, and temporal resolution. To overcome these challenges, the paper introduces a novel event-based autofocus framework that includes an event-specific focus measure called event rate and a robust search strategy known as event-based golden search. In order to achieve better restoration and reconstruction, the event data must be aligned to RGB/intensity sensor data. Therefore, Gu et al.~\cite{gu2021spatio} introduce a novel model for event camera data alignment. This method is particularly effective for extracting camera rotation, leading to improved event alignment. 

EventGPT~\cite{Liu_2025_CVPR} is the first multimodal large language model for event stream understanding. It integrates the event-driven vision into the large language model. The EventGPT contains an event encoder, a spatio-temporal aggregator, a linear projector, an event-language adapter, and an LLM. To overcome the domain gap between LLMs and event information, they adopt three-stage strategies. At the first stage, GPT-generated RGB image-text pairs warm up the linear projector. The second stage uses the NImageNet-Chat dataset, a large synthetic dataset of event data and corresponding texts, to train the spatio-temporal aggregator and event-language adapter. At the final stage, Event-Chat, which contains extensive real-world data, is used to fine-tune the entire model and enhance its generalization ability. Each stage reduces the domain gap between events and LLMs to train EventGPT.

To explore Vulnerabilities in event data processing systems, Wang et al.~\cite{wang2024event} explore the potential risks of backdoor attacks in event-based vision tasks, which have been under-researched despite the increasing use of asynchronous event data. 
A backdoor attack is a method of bypassing normal authentication procedures to gain unauthorized access to a system. This type of cyber attack involves exploiting system vulnerabilities or installing malicious software that creates an entry point for the attacker.
These event-based vision systems can be compromised by injecting malicious triggers into the event data streams, which can then activate the backdoor during inference, leading to incorrect or manipulated outputs.
The authors propose the Event Trojan framework, which includes two types of triggers: immutable and mutable. These triggers are based on sequences of simulated event spikes that can be easily incorporated into any event stream to initiate backdoor attacks. The immutable trigger remains constant, while the mutable trigger uses an adaptive learning mechanism to maximize its aggressiveness. To enhance stealthiness, a novel loss function is introduced to minimize the difference between the triggers and original events while maintaining their effectiveness.

%% file: 30_temporal_domain.tex

\begin{figure*}[!t]
    \centering
    \includegraphics[width=0.80\textwidth]{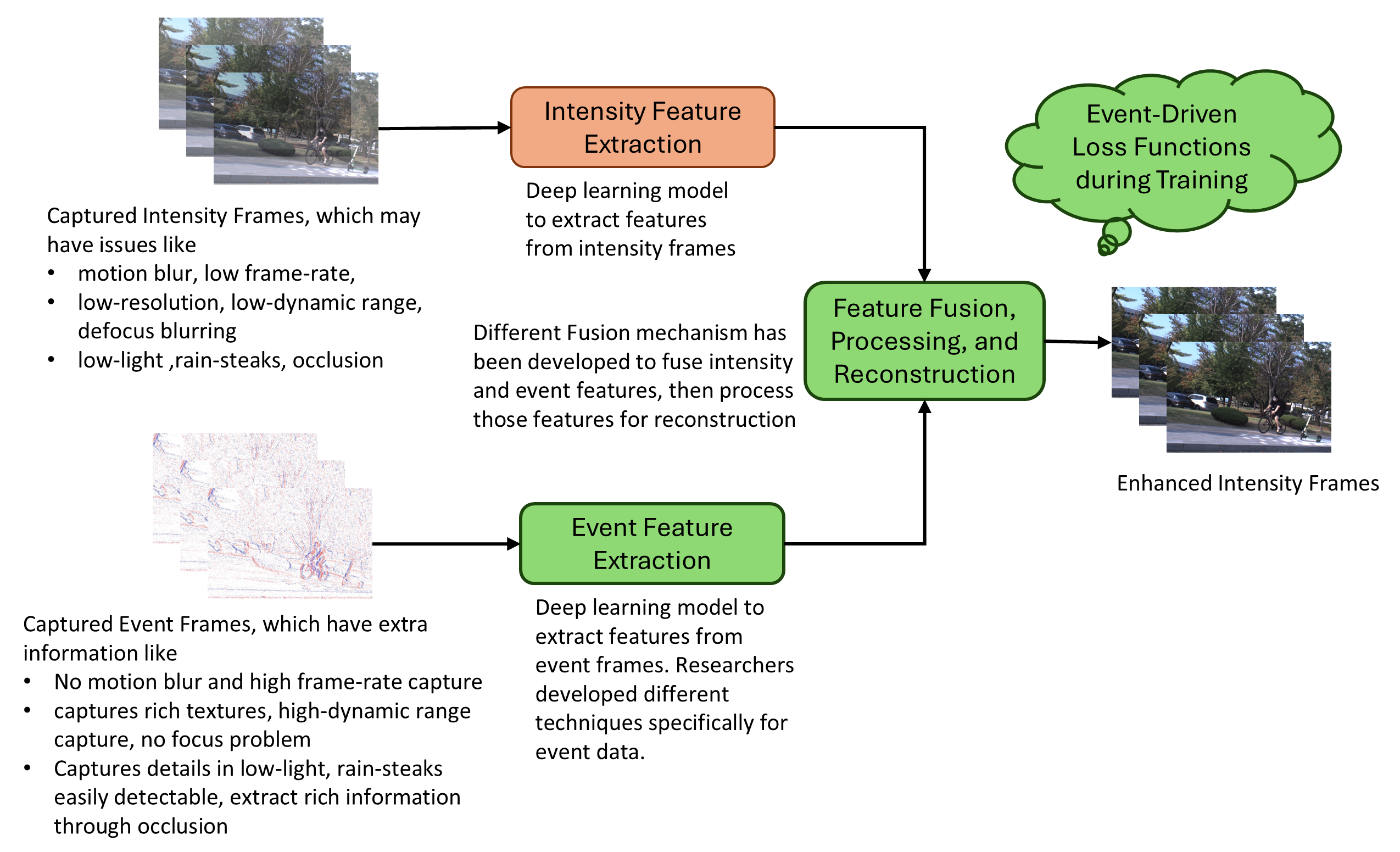}
    \caption{The typical pipeline of how temporal (in Section~\ref{sec:temporal}) and spatial (in Section~\ref{sec:spatial_enhance}) enhancement methods work with event information to restore the degraded images/videos.}
    \label{fig:restore}
\end{figure*}

\section{Temporal Enhancement}  \label{sec:temporal}

\input{30_1_event2video.tex}

\input{30_2_frame_interp.tex}

%% file: 30_1_event2video.tex
\subsection{Event Stream Aided Video Reconstruction}   
\label{subsec:video_recon}

The event-to-video (E2V) problem is a crucial area of research that focuses on reconstructing traditional intensity images or videos from the asynchronous event streams captured by event cameras~\cite{wang2019event, wang2024revisit}. This reconstruction is vital for bridging the gap between event-based vision and conventional frame-based computer vision algorithms. While event cameras offer significant advantages over traditional cameras, the asynchronous structure of output makes directly applying existing computer vision algorithms to event datastreams challenging~\cite{rebecq2019events}. Early methods emphasized the representational similarities between events and gradients~\cite{kim2008simultaneous, cook2011interacting, munda2018real} or optical flow~\cite{bardow2016simultaneous}. However, these approaches often fell short of achieving realistic reconstructions due to a lack of long-term data and prior information exploration. More recently, deep learning techniques, particularly recurrent neural networks (RNNs) and convolutional neural networks (CNNs), have been applied to leverage temporal context for improved performance.

\noindent
\textbf{Handcrafted Priors and Model Based Approaches.}
Early attempts focus on visually interpreting or reconstructing intensity images from pure events. Cook et al.\cite{cook2011interacting} propose utilising recurrently interconnected areas, or maps, to interpret intensity and optic flow. Kim et al.~\cite{kim2008simultaneous} work with pure events on rotation-only scenes to track the camera and build super-resolution mosaics based on probabilistic filtering. This is later extended to 3D Reconstruction and 6-DoF camera motion~\cite{kim2016realtime}. Bardow et al.~\cite{bardow2016simultaneous} reconstruct intensity images and motion fields for generic motion using a variational optimization framework. While applicable to dynamic scenes, these methods often require hand-crafted regularisers, which could lead to a loss of detail in reconstructions. A variational denoising framework is proposed by Munda et al.~\cite{munda2018real}, and it iteratively filters incoming events based on their timestamps to reconstruct images. However, these direct event integration methods often suffer from "bleeding edges" due to non-uniform contrast thresholds and "ghosting effects" from the unknown initial image. Scheerlinck et al.\cite{scheerlinck2018continuoustime} propose a computationally efficient asynchronous filter that continuously fuses image frames and events into a single high-temporal-resolution and high dynamic range image state.

\noindent
\textbf{CNN/RNN Based Approaches.}
The advent of deep learning significantly advances E2V reconstruction, moving beyond restrictive assumptions. These methods typically group events into spatio-temporal representations like 3D voxel grids or event images to be processed by convolutional neural networks (CNNs)~\cite{rebecq2019events, stoffregen2020reducing}. Mostafavi et al.~\cite{mostafavi2018event} utilised conditional Generative Adversarial Networks (GANs) to generate High Dynamic Range (HDR) images and very high frame rate videos from pure events. E2VID~\cite{rebecq2019events} proposes a novel recurrent network architecture (U-Net-based) to reconstruct videos from a stream of events. They train it on large synthetic datasets and show remarkable generalisation to real events, outperforming state-of-the-art reconstruction methods by a large margin in terms of image quality. E2VID+~\cite{stoffregen2020reducing} improves performance by better matching synthetic training data statistics to real-world data. SPADE-E2VID~\cite{cadena2021spadee2vid} introduces spatially-adaptive denormalization (SPADE) layers into the E2VID architecture, which improves the quality of early reconstructed frames but increases computational cost. Zou et al.\cite{zou2021learning} focus on reconstructing high-speed and high-dynamic range videos, and notably collected a paired event and image dataset using a coaxial imaging system, though it was not suitable for high-quality nighttime ground truth generation. HyperE2VID~\cite{ercan2024hypere2vid} proposes a dynamic neural network architecture for event-based video reconstruction that uses hypernetworks to generate per-pixel adaptive filters guided by a context fusion module that combines information from event voxel grids and previously reconstructed intensity images.

\noindent
\textbf{Transformer Based Networks.}
ET-Net~\cite{weng2021event} is the first to explore the application of Transformers for high-speed video reconstruction from an event camera. They propose a hybrid CNN-Transformer architecture (ET-Net) that leverages both fine local information from CNN features and global contexts from Transformers, achieving superior reconstruction quality. Further, a Token Pyramid Aggregation strategy to implement multi-scale token integration for relating internal and intersected semantic concepts in the token-space is also introduced.

\noindent
\textbf{Spiking Neural Networks.}
EVSNN/PA-EVSNN~\cite{zhu2022event} is the first to propose image reconstruction using a deep Spiking Neural Networks (SNNs) architecture. They propose the Event-based Video reconstruction framework based on a fully Spiking Neural Network (EVSNN) and a hybrid potential-assisted framework (PA-EVSNN) utilising Leaky-Integrate-and-Fire (LIF) and Membrane Potential (MP) neurons. Their work aims for greater computational efficiency on event-driven hardware. Tang et al.~\cite{tang2025spike} propose a spike-temporal latent representation (STLR) model for SNN-based E2V reconstruction. This model uses cascaded SNNs, including a Spike-based Voxel Temporal Encoder (SVT) and a U-shape SNN Decoder, to solve the temporal latent coding of event voxels for video frame reconstruction, focusing on energy efficiency.

\noindent
\textbf{Self-Supervised Learning (SSL) Methods.}
Paredesvalls et al.~\cite{paredesvalls2020back} introduce a self-supervised learning framework that estimates optical flow and reconstructs intensity images via photometric constancy. This approach aims to eliminate the dependency on labelled synthetic data, though it still relies on optical flow estimation, which can be prone to errors. EvINR~\cite{wang2024revisit} proposes a novel self-supervised learning approach that re-frames E2V reconstruction as directly solving the event generation model, which is described as a partial differential equation. They utilise Implicit Neural Representations (INRs) \cite{chen2021learning,sitzmann2020implicit} to predict intensity values from spatiotemporal coordinates, eliminating the need for labelled data or optical flow estimation and exhibiting high noise tolerance.

\noindent
\textbf{Diffusion Models.}
Liang et al.~\cite{liang2024e2vidiff} pioneer the use of diffusion models for color video reconstruction from achromatic events. This method addresses the ill-posed nature of E2V (one-to-many mapping) and the regression-to-mean issue by sampling various plausible reconstructions. It employs pretrained diffusion models, a spatiotemporally factorised event encoder, and an event-guided sampling mechanism to ensure faithfulness to original events. Zhu et al.~\cite{zhu2025temporal} introduce a novel framework that leverages temporal and frequency-based event priors to guide a Denoising Diffusion Probabilistic Model (DDPM)~\cite{ho2020denoising}. It uses a temporal domain residual image as the target for the diffusion model and incorporates three conditioning modules: low-frequency intensity estimation, temporal recurrent encoder, and attention-based high-frequency prior enhancement. This approach aims to mitigate over-smoothing and blurry artifacts.

\noindent
\textbf{Other Approaches.}
Cao et al.~\cite{cao2024noise2image} focus on static scene recovery from event cameras by uniquely leveraging illuminance-dependent noise characteristics (photon noise events). It treats noise events as a signal to recover static scene intensity, a component otherwise invisible to event cameras, and can complement E2V methods for scenes with both static and dynamic components. EvTemMap~\cite{bao2024temporal} achieves event-to-dense intensity image conversion using a stationary event camera in static scenes with a transmittance adjustment device (AT-DVS). This method measures the time of event emission for each pixel to form a "Temporal Matrix", which is then converted to an intensity frame using a neural network, aiming to capture diverse scenes beyond dynamic ones. NER-Net~\cite{liu2024seeing} specifically addresses nighttime dynamic imaging, dealing with temporal trailing characteristics and spatial non-stationary distributions of events in low-light conditions. They introduce a Learnable Event Timestamps Calibration module (LETC) and a Non-uniform Illumination Aware Module (NIAM) for this purpose. EventMamba \cite{ge2025eventmamba} observes that the local event relationship in the spatio-temporal domain is key for video restoration tasks and proposes a method comprising Random Window Offset (RWO) in the spatial domain and a new consistent traversal serialization approach in the spatio-temporal domain, which enhances Mamba architecture's capability to handle event data stream.

To summarize, event-to-video reconstruction focuses on creating traditional intensity videos from asynchronous event camera streams, bridging the gap between neuromorphic sensing and conventional vision by leveraging event cameras' high temporal resolution, dynamic range, and lack of motion blur. Approaches have evolved from early hand-crafted or model-based methods to advanced deep learning paradigms, encompassing CNNs, RNNs, Transformers, Spiking Neural Networks, self-supervised learning, and diffusion models, each striving for enhanced reconstruction quality and computational efficiency across various scene conditions.

%% file: 30_2_frame_interp.tex
\subsection{Event-Enhanced Video Interpolation and Deblurring} 
\label{subsec:recon_interp}

Video processing tasks such as video frame interpolation (VFI) and motion deblurring (MD) are crucial for enhancing the quality and temporal resolution of digital videos. Traditional frame-based cameras face inherent limitations, including motion blur due to long exposure times and the loss of inter-frame information caused by slow shutter speeds. These limitations often lead to degraded video quality, especially in dynamic scenes or low-light conditions. Event cameras offer a complementary solution to this challenge. Their ability to capture continuous, non-redundant information about local brightness changes makes them highly advantageous for recovering precise motion cues and filling temporal gaps that conventional cameras miss.

\subsubsection{Video Frame Interpolation}
Video frame interpolation aims to synthesise non-existent intermediate frames between consecutive frames, thereby increasing the video's frame rate. Many VFI methods traditionally rely on linear motion assumptions, which often fail in complex real-world scenarios. Event cameras offer a powerful alternative by providing dense temporal information to model non-linear motions more accurately.

\noindent
\textbf{Flow Based Approaches.}
Flow-based methods typically estimate optical flow between consecutive frames and use this flow to warp input frames to generate intermediate ones.
Flow-based approaches can be broadly classified into two categories: (i) Flow estimation from only the event stream, and (ii) Flow estimation from both the event and the RGB stream.

\begin{itemize}
    \item \emph{Flow from only event stream.} TimeLens~\cite{tulyakov2021time} proposes a learning-based framework that combines warping-based and synthesis-based interpolation. Its warping module estimates optical flow from event sequences, allowing it to handle motion blur and non-linear motion, unlike traditional methods that compute flow from frames and assume linear motion. TimeLens++~\cite{tulyakov2022time} is an extension of TimeLens, improving efficiency and performance by encoding optical flow as cubic splines and using multi-scale feature fusion. It computes a parametric motion model from boundary frames and inter-frame events. TimeReplayer~\cite{he2022timereplayer} introduces an unsupervised learning framework for event-based video interpolation. It directly estimates optical flows between an intermediate frame and input frames using event streams, which helps break the uniform linear motion assumption commonly found in frame-only methods. $A^2OF$~\cite{wu2022video} proposes an end-to-end training method that uses events to generate optical flow distribution masks. These masks provide anisotropic weights for blending and generating intermediate optical flow in orthogonal directions, allowing for a better description of complex motion than isotropic methods. 
    \item \emph{Flow estimation from both the event as well as the RGB stream.}
EIF-BiOFNet~\cite{kim2023event} emphasizes direct estimation of asymmetric inter-frame motion fields by effectively leveraging the distinct characteristics of both events and images, avoiding reliance on approximation methods. It also incorporates an interactive attention-based frame synthesis network. IDO-VFI~\cite{shi2023idovfi} estimates optical flow using both frames and events. It then employs a Gumbel gating module to dynamically decide whether to compute additional residual optical flow in specific sub-regions based on the optical flow amplitude, aiming to reduce computational overhead while maintaining high quality. Chen et al.\cite{chen2023revisiting} suggest that event sequences are better utilized to calibrate optical flow estimated from RGB images rather than for direct optical flow estimation from events alone. They propose an event-guided recurrent warping strategy and a proxy-guided synthesis strategy to better exploit the quasi-continuous nature of event signals. Ma et al.~\cite{ma2024timelensxl} further advance event-based VFI by estimating nonlinear per-pixel motion trajectories between two input frames, enabling it to model more complex motions than previous approaches. It achieves this through iterative motion estimation. TimeTracker~\cite{liu2025timetracker} proposes a continuous point tracking method consisting of a Scene Aware Region Segmentation (SARS) module and a Continuous Trajectory guided Motion Estimation (CTME) module, which is finally used in global motion optimization followed by frame refinement. The experiments show the superiority of TimeTracker in fast nonlinear motion scenarios.
\end{itemize}

\noindent
\textbf{Synthesis Based Approaches.}
Synthesis-based methods directly regress intermediate frames or fuse information without explicitly relying on optical flow estimation or image warping operations. Liu et al.~\cite{liu2024video} propose a purely synthesis-based E-VFI framework. It directly synthesizes interpolated frames by aligning keyframes to an event-based reference that encodes structural information, thereby circumventing the challenges of non-linear motion fitting and occlusion issues typically introduced by warping operations. Gao et al.~\cite{gao2022superfast} introduce a fast-slow joint synthesis framework for high-speed VFI. It divides the task into two sub-tasks focusing on contents with and without high-speed motions, tackled by a fast synthesis pathway (including an SNN-based hybrid module for high-speed content) and a slow synthesis pathway, respectively. RE-VDM~\cite{chen2025repurposing} proposes to adapt pretrained video diffusion models for frame interpolation to overcome the data scarcity problem. By using Per-tile Denoising and Two-side fusion, it achieves better interpolation consistency and generalization on unseen real-world data. To tackle the problem of latency in interpolating between two frames, Musunuri et. al \cite{musunuri2024event} propose to extrapolate RGB frames from the initial RGB frame and asynchronous events.

\noindent
\textbf{Hybrid and Attention Based Approaches.}
Hybrid approaches combine elements of both flow-based and synthesis-based approaches, or heavily rely on attention mechanisms and feature fusion for robust interpolation. Kiliccet et al.~\cite{kilicc2022evfia} propose a lightweight, kernel-based method that fuses event information with standard video frames using deformable convolutions and a multi-head self-attention mechanism. This approach aims to generate crispier frames less vulnerable to blurring and ghosting artifacts. Akmanet et al.~\cite{akman2023maevima} introduce motion awareness through event-based motion masks, which guide deformable convolutions during image generation to focus on moving areas. It also incorporates a motion-aware loss function to further improve performance. Zhanget et al.~\cite{zhang2023extracting} propose a unified operation using inter-frame attention to explicitly extract both motion and appearance information simultaneously. It reuses the attention map for both appearance feature enhancement and motion information extraction, aiming for efficiency. Jin et al.~\cite{jin2022aup} follow a flow-guided synthesis framework, employing optical flow estimation, warping of input frames and context features, and synthesizing intermediate frames from these warped representations. Nottebaum et al.~\cite{nottebaum2022efficient} propose a method that fine-tunes an initial block-based Principal Component Analysis (PCA) basis end-to-end for video frame interpolation, creating a general projection space for all images and resolutions and optimizing the representation for the VFI task. Wang et al.~\cite{wang2023eventbased} propose a novel task of event-based continuous color video decompression. It employs a joint synthesis and motion estimation pipeline, where the synthesis module uses a K-plane-based factorization to encode event-based spatiotemporal features, and the motion estimation module estimates time-continuous nonlinear trajectories. Cho et al.~\cite{cho2024tta} introduce a Test-Time Adaptation (TTA) framework for event-based VFI to address the significant performance drop experienced by event-based VFI methods when applied to different domains (i.e., test datasets with varying distributions from training data). They achieve this by using confident pixels as pseudo ground-truths, while mitigating overfitting to recurring scenes by blending historical samples with current inputs. This is crucial for real-world applications where device and environment variations are unavoidable.

\subsubsection{Motion Deblurring}
Motion deblurring is a challenging and ill-posed problem due to the loss of motion information and intensity textures during the blur degradation process. Event cameras offer a unique advantage by inherently providing precise motion information and sharp edges, which can be leveraged to alleviate motion blur.

\noindent
\textbf{Model Based Approaches.}
Model-based approaches derive deblurring solutions based on the physical principles of event generation and image formation, sometimes coupled with optimization techniques. Pan et al.~\cite{pan2019bringing} propose the Event-based Double Integral (EDI) model to connect intensity images with event data by integrating over events as well as image sequence while considering the blur process. It aims to recover a sharp image and then reconstruct a high frame-rate video, implying a sequential or integrated approach to deblurring and high frame-rate reconstruction. Pan et al.~\cite{pan2019high} extend EDI to multiple Event Based Double Integral (mEDI) to analytically reconstruct high frame-rate sharp videos from blurry frames and associated events. This work specifically highlights the challenge of addressing blur in combined event-intensity solutions. E-CIR~\cite{song2022ecir} extends the EDI model by using a deep learning model to predict a sharp video represented by parametric polynomials from a blurry frame and associated events. It also formulates a refinement objective to encourage temporal propagation of sharp visual features and address temporal smoothness issues.

\noindent
\textbf{Learning Based Approaches.}
Learning-based methods employ neural networks (CNNs, RNNs, attention mechanisms, etc.) to learn the mapping from blurry inputs and event data to sharp outputs, often suppressing noise and handling complex real-world conditions. Lin et al.~\cite{lin2020learning} employ a network to estimate intensity residuals between latent sharp and blurry images directly from event integrals, effectively using dynamic filters to handle spatially and temporally variant triggering thresholds. Sun et al.~\cite{sun2022event} propose an Event-Image Cross-modal Attention fusion module (EFNet) that jointly extracts and fuses information from event streams and images. They demonstrate that the principled architecture leverages event information more effectively for deblurring compared to simple concatenation or optical flow estimation. Kim et al.~\cite{kim2021eventguided} propose a novel Exposure Time-based Event Selection(ETES) module to selectively use event features by estimating the cross-modal correlation between the features from blurred frames and the events, and propose a feature fusion module to fuse the selected features from events and blur frames effectively. Sun et al.~\cite{sun2024motion} introduce Deviation Accumulation (DA) as an event preprocessing method to enhance motion perception, enabling the model to distinguish different motion patterns. DA considers the average deviation of polarities at each pixel position over the exposure period. The model also features a Recurrent Motion Extraction (RME) module for multi-scale motion extraction and a Feature Alignment and Fusion (FAF) module to mitigate inter-modal inconsistencies. Kim et al.~\cite{kim2024cmta} focus on exploiting long-range temporal dependencies in videos. They propose intra-frame feature enhancement through recurrent cross-modal interactions within the exposure time and inter-frame temporal feature alignment to gather information from surrounding adjacent frames. DiffEvent~\cite{wang2024diffevent} proposes to formulate event-based image deblurring as an image generation problem using diffusion priors for the image and residual, introducing an alternative diffusion sampling framework. CrossZoom~\cite{zhang2023crosszoom} introduces a novel unified neural Network (CZ-Net) to jointly recover sharp latent sequences within the exposure period of a blurry input and the corresponding High-Resolution (HR) events and presents a multi-scale blur-event fusion architecture that leverages the scale-variant properties and effectively fuses cross-modal information to achieve cross-enhancement. Zhang et al.\cite{zhang2023neuromorphic} introduce a unified processing structure for joint restoration of blurry images and noisy events. It proposes an event-regularized prior for blind deblurring and uses gradient priors from recovered sharp images to supervise event denoising, aiming for robust performance under severe blur and noise. SC-Net~\cite{cao2023event} proposes a hybrid Spiking Neural Network (SNN) and Convolutional Neural Network (CNN) architecture for various video restoration tasks, including deblurring. It features a Spiking Neural Temporal Memory (SNTM) for long-term temporal event correlation and a Frame-Event Spatial Aggregation (FESA) module for spatial consistency. Zhang et al.~\cite{zhang2023generalizing} propose a Scale-Aware Network (SAN) with a Multi-Scale Feature Fusion (MSFF) module for spatially continuous representation and an Exposure-Guided Event Representation (EGER) for arbitrary target latent images, along with a two-stage self-supervised learning framework to generalize deblurring performance across spatial and temporal domains. NED-Net~\cite{cho2023noncoaxial} specifically tackles deblurring with non-coaxial event cameras. It proposes an Attention-based Deformable Align (ADA) module for robust feature-level alignment between image and event data, a Local Score-based Aggregation (LSA) module, and a Cross-Channel Interaction (CCI) module for texture enhancement. St-EDNet~\cite{lin2023learning} proposes a coarse-to-fine framework to effectively utilize and aggregate information from a single blurry image and corresponding event streams (even if misaligned) from a stereo setup. It simultaneously outputs a sequence of sharp images and a disparity map, leveraging parallax to mitigate artifacts caused by misalignment. EGDeblurring~\cite{xie2025diffusion} uses a diffusion model to generate event guidance, which then can be used to deblur RGB images. This enables the usage of event-based deblurring even in cases with RGB-only capture.

\subsubsection{Unified Deblurring and Frame Interpolation}
This class of methods explicitly addresses both deblurring and frame interpolation as a unified task, or those whose primary goal of high frame-rate video reconstruction from blurry inputs inherently requires both. Lin et al.~\cite{lin2020learning} explicitly state their aim to reconstruct a sharp video with an increased frame rate from low frame-rate blurry videos and corresponding event streams. The proposed algorithm consists of residual estimation, keyframe deblurring, and video frame interpolation components. Zhang et al.~\cite{zhang2022unifying} present a unified framework for event-based motion deblurring and frame interpolation for blurry video enhancement. They employ a learnable double integral (LDI) network and a fusion network, and crucially, a fully self-supervised learning framework that enables training with real-world blurry videos and events without ground truth images. The LDI network is designed to automatically predict the mapping relation between blurry frames and sharp latent images from the corresponding events. Zhang et al.~\cite{zhang2023neural} propose a unified neural network framework that can "re-expose" a captured photo by adjusting its "neural shutter". This allows it to handle multiple shutter-related tasks simultaneously, including image deblurring, video frame interpolation, and rolling shutter correction. Cheng et al.~\cite{cheng2023recovering} focus on the task of continuous-time video extraction from a single blurry image using events. This inherently involves both deblurring the input image and interpolating a high frame-rate sequence, enabling restoration of sharp latent images at arbitrary timestamps. Lu et al.~\cite{lu2023selfsupervised} propose a novel self-supervised framework that leverages events to guide Rolling Shutter (RS) frame correction and VFI in a unified manner. It aims to recover arbitrary frame rate Global Shutter (GS) frames from two consecutive RS frames. Sun et al.~\cite{sun2023event} present a general method for event-based frame interpolation that performs deblurring ad hoc, making it applicable to both sharp and blurry input videos. It uses a bidirectional recurrent network and an event-guided channel-level attention fusion module. Weng et al.~\cite{weng2023eventbased} study the challenging problem of blurry frame interpolation under blind exposure (i.e., unknown and dynamically varying exposure time) with event cameras. They propose an exposure estimation strategy guided by event streams and a temporal-exposure control strategy for arbitrary-time interpolation. Yang et al.~\cite{yang2024latency} develop the first method to explicitly address latency correction in improving event-guided deblurring and interpolation tasks. Event timestamps often deviate from actual intensity changes due to sensor latency, noise, and illumination variability, making the temporal discrepancy spatially inconsistent and difficult to model. They introduce an event-based temporal fidelity metric to evaluate the sharpness of reconstructed images from latency-corrected events. Lu et al.~\cite{lu2023uniinr} pioneer the simultaneous exploration of RS correction, deblurring, and VFI within a one-stage framework using a unified implicit neural representation (INR). This approach boasts high efficiency and a lightweight model.

To summarize, this section outlines a robust and evolving landscape for event-based video restoration, moving from initial efforts to recover high frame-rate videos to sophisticated unified frameworks addressing multiple image degradation issues simultaneously. While early works focused on either deblurring or interpolation, a clear trend towards jointly addressing motion deblurring and video frame interpolation has emerged, often integrated with other challenging tasks like rolling shutter correction and continuous intensity recovery. Future advancements will likely continue to refine cross-modal fusion techniques, improve generalization to real-world data, and develop more efficient and lightweight architectures to enable widespread practical applications.

%% file: 40_spatial_domain.tex

\section{Spatial Enhancement}   
\label{sec:spatial_enhance}

The event camera excels at capturing scene changes in high dynamic range scenarios at an exceptionally high frame rate without motion blur. These distinctive features of event camera frames contribute significantly to spatial video enhancement. Various spatial video enhancements can be achieved by fusing event camera frames with conventional RGB camera frames. This section will cover topics such as super-resolution and artifact reduction, HDR enhancement, low-light enhancement, occlusion removal, rain removal, and focus control. Literature reviews indicate that such spatial enhancements are feasible through the fusion of event camera information.

\subsection{Super-Resolution and Artifact Reduction}    
\label{subsec:super_res}

Event cameras capture small spatial textural details in the event stream and help achieve video super-resolution and artifact reduction. Different literature shows unique approaches to fuse the event information along with low-resolution RGB/ intensity frames to reconstruct high-quality output frames. Those approaches can be classified into three different categories, namely Event Information Fusion, Event Information Alignment for Fusion, and Event Information Representation for Fusion.

\noindent
\textbf{Event Information Fusion.} 
The algorithms discussed in this paragraph take both low-resolution images and event data as input to develop a fusion technology-based novel architecture for super-resolution. Those approaches assume the alignment between events and intensity images.
Wang et al.~\cite{wang2020event} propose an event-enhanced sparse learning network (eSL-Net), which takes a low-resolution image and an event pair as input and predicts the high-resolution image. 
The sparse learning network is based on the idea of compressed sensing and dictionary learning, where both Low-Resolution (LR) and High-Resolution (HR) images are assumed to have sparse representations in learned dictionaries.
eSL-Net is designed by unfolding the Iterative Shrinkage Thresholding Algorithm (ISTA) into a deep neural network. This makes the network interpretable, as each layer corresponds to an iteration of ISTA.
ISTA is a widely used optimization method for solving sparse coding problems, especially in contexts like compressed sensing and image reconstruction.
In a single framework, eSL-Net performs deblurring and denoising together while performing super-resolution. 
Jing et al.~\cite{jing2021turning} uncover a pivotal insight in video super-resolution. They discover regions with higher temporal frequency—characterized by smaller pixel displacements between consecutive frames—contribute more effectively to the reconstruction of high-resolution textures. These tiny displacements are crucial because they preserve fine-grained details and subtle variations in the scene, which are often lost in lower-frequency regions with larger motion gaps.
This observation leads to a paradigm shift in video super-resolution design and motivates the use of an event camera for video super-resolution, as an event camera captures uniform and tiny pixel displacements between neighboring frames.
These cameras produce dense streams of events that inherently encode high-frequency motion information, making them ideal for capturing texture-rich regions.
They propose an asynchronous interpolation (EAI) module in their video super-resolution (VSR) framework, which effectively fuses event and image features. 
EvIntSR-Net~\cite{han2021evintsr} converts event data into multiple latent intensity frames to reduce the domain gap between event streams and intensity frames. As latent domain extracts high-level information, it is relatively easy to bring the event stream latent closer in the latent domain of intensity frames.
Kai et al.~\cite{kai2023video} introduce an Event-driven Bidirectional Video Super-Resolution (EBVSR) framework, which has an event-assisted temporal alignment module that leverages events to generate nonlinear motion for aligning adjacent frames. 
Guo et al.~\cite{guo2023event} develop EFSR-Net, which contains coupled response blocks (CRB) that fuse data from both RGB and Event camera. This enables the recovery of detailed textures in shadows. 
Noise present in event streams hinders the performance of super-resolution. Separate additional event denoising may lead to over-suppression of events. Therefore, Yu et al.~\cite{yu2023learning} develop an event-enhanced Sparse Learning Network (eSL-Net++) based on a dual sparse learning scheme, where both events and intensity frames are modeled with sparse representations to suppress the noise present in the event stream and handle low-resolution with motion blurs together. An event shuffle-and-merge scheme is also proposed to extend eSL-Net++ to generate high-resolution and high-framerate video from a single blurry frame without additional training. 
Liu et al.~\cite{liu2024disentangled} propose an end-to-end framework called EGI-SR, which employs three Cross-Modality Encoders to learn both modality-specific and modality-shared features from stacked events and intensity images. This design helps mitigate the negative impact of modality differences and reduces the feature space gap between events and intensity images. A transformer-based decoder is then used to reconstruct the super-resolved image. Unlike traditional VSR methods that focus on motion learning, EvTexture~\cite{kai2024evtexture} utilizes the high-frequency details captured by event cameras to improve texture regions. 
MamEVSR~\cite{Xiao_2025_CVPR}, a Mamba-based deep neural network~\cite{gu2021efficiently, gu2023mamba} for event guided video super-resolution. Their mamba-driven network offers a global receptive field with linear computational complexity. By doing so, it overcomes the limitations of the convolutional neural network and transformers. MamEVSR's interleaved Mamba (iMamba) block applies multidirectional selective state space modeling for feature fusion and propagation across bi-directional frames while maintaining linear complexity. MamEVSR's cross-modality Mamba (cMamba) block exploits spatio-temporal information from both event information and the output of the iMamba block and fuses the cross-modality information.
Unlike other approaches, which perform super-resolution for a specific scale factor, Lu et al.~\cite{lu2023learning} reconstruct high-resolution frames with an arbitrary scale factor. Their framework consists of the Spatial-Temporal Fusion (STF) module to learn 3D features from both events and RGB frames, the Temporal Filter (TF) module to extract explicit motion information from events near the queried timestamp to generate 2D features, and the Spatial-Temporal Implicit Representation (STIR) module for arbitrary scale super-resolution.

\noindent
\textbf{Event Information Alignment for Fusion.}
The algorithms discussed in this paragraph deal with the alignment problem between events and intensity images, which is a real-world challenge and is crucial for super-resolution. The existing methods discussed above assume strict calibration between event and RGB cameras, which is often impractical for high-resolution devices like dual-lens smartphones and drones. Here, we will discuss the algorithms that develop a method to align event information with intensity information to perform super-resolution.
Asymmetric Event-guided Video Super-Resolution Network (AsEVSRN)~\cite{xiao2024asymmetric} addresses this by incorporating two specialized designs: the content hallucination module and event-enhanced bidirectional recurrent cells. The content hallucination module dynamically enhances event and RGB information, while the bidirectional recurrent cells align and propagate temporal features using event-enhanced flow. 
In a similar direction, Xiao et al.~\cite{xiao2024event} introduce the Event AdapTER (EATER), which includes the event-adapted alignment (EAA) unit and the event-adapted fusion (EAF) unit. The EAA unit aligns multiple frames using event streams in a coarse-to-fine manner, while the EAF unit fuses these frames with event data through a multi-scale design.

\noindent
\textbf{Event Information Representation for Fusion.} 
Event representation and event feature extraction also play a crucial role in super-resolution. The methods discussed below presented different techniques to represent or extract event information that benefits super-resolution performance.  
Teng et al.~\cite{teng2022nest} introduce a novel event representation called Neural Event Stack (NEST), which encodes comprehensive motion and temporal information while adhering to physical constraints and suppresses noises in the event stream, making it suitable for image enhancement tasks like super-resolution and deblurring. 
Event-based Blurry Super Resolution Network (EBSR-Net)~\cite{zhang2024super} includes a multi-scale center-surround event representation to exploit intra-frame motion to extract multiscale texture information inherent in events. 
For better event feature extraction, Cao et al.~\cite{cao2023event} develop a spiking-convolutional neural network (SC-Net), which integrates spiking neural networks (SNNs) with convolutional neural networks (CNNs) to process asynchronous event data. The proposed method includes a spiking-convolutional layer that extracts temporal features from event streams. Apart from super-resolution, it performs deblurring and deraining.

\subsection{HDR Enhancement}    
\label{subsec:hdr_enhancement}

Traditional High Dynamic Range (HDR) imaging techniques often involve merging multiple Low Dynamic Range (LDR) images taken at different exposures. This can be challenging due to over- or under-exposure issues, leading to ghosting artifacts. On the other hand, event cameras capture high dynamic range scenes as intensity maps. The incorporation of LDR RGB frames and event fusion offers a promising solution for capturing high-quality HDR images in diverse lighting conditions.
Literature shows multiple novel HDR imaging pipelines~\cite{han2020neuromorphic, messikommer2022multi, han2023hybrid, shaw2022hdr, yang2023learning, weng2024event, li2024generalizing} that integrate bracketed LDR images data from standard cameras and HDR events from event cameras to reconstruct HDR images.
Shaw et al.~\cite{shaw2022hdr} develop an event-to-image feature distillation module that translates event features into the image-feature space with self-supervision and use attention and multi-scale spatial alignment modules for fusion. 
HDRev-Net~\cite{yang2023learning} utilizes temporal correlations recurrently to suppress flickering effects in the reconstructed HDR video.   
Weng et al.~\cite{weng2024event} propose a lightweight multi-scale receptive field block that is employed for rapid modality conversion from event streams to frames, while a dual-branch fusion module aligns features and removes ghosting artifacts caused by differences in camera positions and frame rates. 
An exposure-aware framework~\cite{li2024generalizing} includes an exposure attention fusion module and incorporates a self-supervised loss based on structural priors to enhance details in saturated areas and reduce noise. 
Exposure attention fusion is a mechanism designed to adaptively fuse features from SDR images and event streams based on the exposure level of different regions in the image. It uses an exposure mask to guide this fusion process.
Guo et al.~\cite{guo2024event} introduce a diffusion-based fusion module and a real-world finetuning strategy to enhance the generalization of the alignment module on real-world events. They incorporate image priors from pre-trained diffusion models to address artifacts in high-contrast regions and minimize alignment errors.
Color events record asynchronous pixel-wise color changes in a high dynamic range. Cui et al.~\cite{cui2024color} incorporate color events into the single-exposure HDR imaging pipeline.  An exposure-aware transformer (EaT) module is designed to propagate informative hints from normally exposed LDR regions and event streams to the missing areas. This module includes an exposure-aware mask, which is a learned attention-like mechanism that helps guide the transformer to focus on reliable regions during HDR reconstruction. It suppresses misleading information from saturated regions and enhances the propagation of trustworthy color hints from well-exposed areas.
ERS-HDRI~\cite{li2024ers} is designed to enhance remote sensing images, which often struggle with low dynamic range, leading to incomplete scene information. The proposed framework addresses this by employing a coarse-to-fine strategy, integrating the event-based dynamic range enhancement (E-DRE) network and the gradient-enhanced HDR reconstruction (G-HDRR) network. The E-DRE network extracts dynamic range features from LDR frames and event streams, performing intra- and cross-attention operations to fuse multi-modal data. A denoise network and a dense feature fusion network generate a coarse HDR image, which is then refined by the G-HDRR network using a gradient enhancement module and a multiscale fusion module.
Self-EHDRI~\cite{xiaopeng2024hdr}, a self-supervised learning paradigm, generalizes HDR enhancement performance in real-world dynamic scenarios. This framework employs a self-supervised learning strategy to learn cross-domain conversions from blurry LDR images to sharp LDR images, enabling the recovery of sharp HDR images even without ground-truth sharp HDR images.
Unlike existing, AsynHDR~\cite{wu2024event} does not combine RGB frame-based sensors with event sensors in the same system.
AsynHDR is a pixel-asynchronous HDR imaging system, a novel capture system that integrates both event-based sensors (DVS) and optical modulation components (LCD panels). It integrates event sensors with LCD panels that modulate the irradiance incident upon the event sensors, triggering pixel-independent event streams.
It achieves HDR imaging solely using DVS, modulated light, and a novel temporal-weighted reconstruction algorithm. The system encodes scene brightness into the timing of events, which enables reconstruction of HDR images with high dynamic range and reduced noise. The authors mention that if a DVS sensor with a Bayer matrix is used, the system could be extended to color HDR imaging, similar to RGB cameras.

\subsection{Lowlight Enhancement}   
\label{subsec:lowlight_enhace}

Traditional RGB cameras often struggle with long exposure times in low-light conditions, leading to motion blur and reduced visibility. With their high dynamic range and temporal resolution, event cameras can effectively capture motion information even in very dark settings. This unique capability of event cameras helps to achieve low-light enhancement. 
Zhang et al.~\cite{zhang2020learning} propose a novel unsupervised domain adaptation network that translates HDR events in low light into sharp, canonical images as if captured in daylight. 
Event cameras naturally provide HDR data, which preserves coarse scene structure even in darkness.
The domain adaptation framework leverages this HDR property to bridge the gap between noisy, sparse, low-light events and rich daylight images without any need for paired data.
Liu et al.~\cite{liu2024seeing} develop a nighttime event to video reconstruction network (NER-Net) that includes a learnable event timestamps calibration module to align temporal trailing events and a non-uniform illumination aware module to stabilize the spatiotemporal distribution of events. The above-mentioned works mainly convert events in low-light conditions into intensity videos to improve the visibility in low-light scenarios. Other works mainly rely on fusion strategies~\cite{liu2023low, wang2024exploring, jiang2023event, yunfansee}. Incorporation of event information with low-light intensity videos is used to elevate the performance of the low-light scene enhancement framework.
To ensure temporal stability and restore details, Wang et al.~\cite{wang2024exploring} design unsupervised temporal consistency loss and detail contrast loss, which, along with supervised loss, contribute to the semi-supervised training of the network on unpaired real data. 
Jiang et al.~\cite{jiang2023event} introduce a residual fusion module to minimize the domain gap between event streams and frames by utilizing the residuals of both modalities. Here, residuals refer to residual connections or residual blocks commonly used in deep learning architectures, especially in CNNs and transformer-based models.
Lu et al.~\cite{yunfansee} propose SEE-Net, a novel framework for enhancing images captured under a wide range of lighting conditions using event camera data. It takes as input a standard RGB image and its corresponding event stream, and outputs a brightness-adjustable image guided by a user-defined brightness prompt. The method begins by embedding spatial and sensor-specific information into both image and event data. These features are fused using cross-attention mechanisms to form a Broader Light Range (BLR) representation, which captures illumination dynamics across lighting extremes. A multi-layer perceptron (MLP) decoder then integrates the brightness prompt to generate the final enhanced image, allowing pixel-level control over exposure. This design enables flexible brightness adjustment during inference and robust training across diverse lighting scenarios.
They also propose SEE-600K, a large-scale dataset comprising $610,126$ image-event pairs collected from 202 distinct scenes, each captured under approximately four different lighting conditions, spanning a dynamic illumination range of over $1,000$-fold.
Liang et al.~\cite{liang2024towards} address the limitations of existing research by introducing a large-scale dataset comprising over $30,000$ pairs of images and events captured under varying illumination conditions. This dataset was meticulously curated using a robotic arm to ensure precise spatial and temporal alignment. The proposed method, EvLight, integrates structural and textural information from both images and events through a multi-scale holistic fusion branch. To handle variations in regional illumination and noise, the authors introduce SNR-guided regional feature selection, enhancing features from high SNR regions and augmenting those from low SNR regions by extracting structural information from events. The extension of the approach, EvLight++, is developed by Chen et al.~\cite{chen2024evlight++}. It extends the method to videos and proposes some modifications, like a ConvGRU recurrent module to capture long-range temporal dependencies, a temporal loss to ensure illumination consistency across frames, and new spatio-temporal alignment using a matching strategy.

Along with low-light enhancement, a few works propose algorithms to handle multiple degradations during low-light scenes.
Liang et al.~\cite{liang2023coherent} effectively address challenges like motion blur and noise in low-light conditions. They use hybrid inputs of events and frames to capture temporal correspondences and provide alternative observations, such as intensity ratios between consecutive frames and exposure-invariant information. A neural network is trained to establish spatiotemporal coherence between visual signals of different modalities and resolutions by constructing a correlation volume across space and time.  
Kim et al.~\cite{kim2024towards} develop an end-to-end framework that leverages temporal information from both events and frames, incorporating a cross-modal feature module to enhance structural information while suppressing noise to perform low-light video enhancement and deblurring. They also develop the Real-world Event-guided Low-light Enhancement and Deblurring (RELED) dataset, which includes synchronized low-light blurred images, normal-light sharp images, and low-light event streams. 
Zhang et al.~\cite{zhang2024sim} address the challenges of video frame interpolation (VFI) in low-light conditions using event cameras and propose a novel per-scene optimization strategy that leverages the internal statistics of a sequence to handle degraded event data. This approach improves the generalizability of VFI to different lighting and camera settings.

\subsection{Occlusion Removal}      
\label{subsec:occlusion_removal}

Traditional synthetic aperture imaging (SAI) methods often require prior information and have strict camera motion constraints. They also struggle with dense occlusions and extreme lighting conditions, leading to performance degradation. Event camera helps to overcome these limitations and provides a robust solution for capturing detailed images with significant occlusions, enhancing the capabilities of imaging systems in various applications.
REDIR~\cite{guo2024redir}, an end-to-end refocus-free variable event-based SAI method. It aligns global and local features of variable event data to achieve occlusion-free imaging from pure event streams without needing prior information (like camera motion parameters or manual focusing). 
REDIR incorporates a Perceptual Mask-Gated Connection Module (PMCM) to interlink information between modules and a Temporal-Spatial Attention (TSA) mechanism within the Spiking Neural Network (SNN) block to enhance target extraction. 
PMCM connects the registration and reconstruction parts of the network. It filters occlusion events and helps transfer useful features between modules.
The TSA mechanism helps the model focus on persistent target signals across time and space, filtering out transient occlusion noise.
The event registration module uses a neural network to align event data from different times.
It handles camera shake, rotation, and zoom automatically.
Occlusion Filtering filters out noise from occluding objects using a smart attention mechanism. It keeps only the useful signals from the target.
Unlike REDIR, other approaches adopt fusion strategies for occlusion removal. Most of the approaches consist of a hybrid network combining spiking neural networks (SNNs) and convolutional neural networks (CNNs). The SNN layers encode the spatio-temporal information from the event data, while the CNN decoder transforms this information into visual images of the occluded targets.
CNNs and SNNs-based fusion systems combine the strengths of event cameras and frame-based cameras to generate occlusion-free visual images in both sparse and dense occlusions~\cite{zhang2021event, liao2022synthetic, li2022image, yu2022learning, guo2024improved}.
The EF-SAI system of ~\cite{liao2022synthetic} processes multi-modal features through a multi-stage fusion network that enhances cross-modal information and selects density-aware features. 
Li et al.~\cite{li2022image} present a comparison loss function that is introduced to enhance the clarity of the reconstructed images.
They introduce an SNN-based encoder that denoises and encodes the asynchronous event data. The encoded event features and frame features are fused using a custom-designed fusion layer, and a joint decoder reconstructs the final clear image.
Zhang et al.~\cite{zhang2023unveiling} capture continuous streams of events from occluded scenes and integrate event information from continuous viewpoints using a novel cross-view mutual attention mechanism for effective fusion and refinement. 
For occlusion removal, the event camera is rapidly moved across a scene. This movement allows the camera to capture the scene from slightly different perspectives over time, and it effectively simulates multiple views of the same scene.
In another work, Zhang et al.~\cite{zhang2024transient} process these events by integrating multi-view spatial-temporal information through a Long-Short Window Feature Extractor (LSW) and a cross-view mutual attention-based module for improved fusion and refinement. 
The LSW is designed to handle the spatial-temporal richness of spike data. 
Dense window representation preserves fine-grained temporal details.
Long window representation accumulates spikes over a longer time to simulate a blurred image-like view, which is helpful to capture structural information.
These representations are then fused using a Cross-View Attention (CVA) module.
Guo et al.~\cite{guo2024improved} present an event stream encoder based on SNNs, which efficiently encodes and denoises the event data for de-occlusion. 
Instead of binary spikes, they used Full-Precision LIF (FP-LIF) neurons, which retain the actual membrane potential values when firing. This prevents information loss during encoding, which is a challenge in traditional LIF. FP-LIF helps to improve feature quality.
They also introduce Isomorphic Network Knowledge Distillation to solve the limited data issue. 
Yu et al.~\cite{yu2022learning} use a Refocus-Net module, which refocuses collected events to align in-focus events while scattering out off-focus ones. 
The concept of refocusing is used to address the problem of occlusion.
The proposed method refocuses event streams to align signal events from occluded targets while scattering noise events from foreground occlusions. This helps to achieve high-quality reconstruction even under very dense occlusions.
Previous methods could only perform at a single depth plane, which limits their usefulness in scenes with multiple depth layers.
Liu et al.~\cite{liu2025all} introduce a method to refocus each event individually based on its depth, and it enables occlusion-free imaging across multiple depths.
They prove that it is feasible to estimate depth maps from multi-view event data even under dense occlusions, which was previously considered very challenging.
They introduce a Depth Estimation Module (DEM) that uses a Time-Guided Attention Module (TGAM) to capture temporal and viewpoint relationships.
Deformable Residual Blocks (DRB) to adaptively extract features across scales.
The entire pipeline is trained end-to-end using only occlusion-free multi-view images as supervision. There is no need for ground-truth depth maps.

\subsection{Rain Removal}   
\label{subsec:rain_removal}

Video deraining models using events contribute to the field by offering a powerful tool for improving video quality in adverse weather conditions. This network leverages the unique capabilities of event cameras to separate rain streaks from the background, enhancing their utility in challenging weather conditions. Experimental results show that these methods outperform existing deraining techniques for traditional camera videos. These approaches represent a significant advancement in the field of video deraining, offering promising potential for future research and applications.
The approach developed by Cheng et al.~\cite{cheng2022novel} operates in the width and time (W-T) space, leveraging the discontinuity of rain streaks in these dimensions while maintaining the smoothness of background objects. 
"width" refers to the horizontal spatial dimension of the event frame, which is the number of pixels across each row of the image.
3D event video data (height $\times$ width $\times$ time) is transformed into a W–T space by slicing along the height axis.
In this W–T space, each image is a 2D slice where the horizontal axis is the width (i.e., pixel columns), and the vertical axis is time (i.e., event frame index).
This transformation helps reveal the discontinuity of rain streaks along the width and time dimensions, as the rain typically moves vertically and affects only a few adjacent pixels in width, it appears as sparse, noise-like patterns in W–T space.
This allows raindrops and streaks to be treated as uniform noise, which can be effectively removed using a non-local means filter.

Later, Zhang et al.~\cite{zhang2023egvd}'s end-to-end learning-based network that includes an event-aware motion detection module, which adaptively aggregates multi-frame motion contexts using event-aware masks to capture motion information from event streams effectively, and a pyramidal adaptive selection module, which reliably separates background and rain layers by incorporating multi-modal contextualized priors, ensuring accurate rain removal and preservation of background details.
As rain steaks can be identified easily in event data, Wang et al.~\cite{wang2023unsupervised} leverage this and introduce an end-to-end unsupervised learning-based network for the first time. It consists of two key modules: the Asymmetric Separation Module to segregate features of the rain and background layers, and the Cross-Modal Fusion Module to enhance positive features and suppress negative ones from a cross-modal perspective. 
To address the difficulty of modeling the temporal and spatial correlations of rain streaks in videos with existing methods, Sun et al.~\cite{sun2023event} propose approach include Multi-Patch Progressive Learning, which divides video frames into multiple patches and processes them progressively to capture intricate rain streak details, and an Event-Aware Mechanism that leverages event-based sensors to detect and remove rain streaks effectively.
Similar to other domains, spiking convolutional neural networks have been adopted in deraining to adapt sparse event sequences to capture the features of falling rain~\cite{fu2024event, ruan2024distill}. Fu et al.~\cite{fu2024event} introduce a Bimodal Feature Fusion Module, which combines dense convolutional features from video frames with sparse spiking features from event sequences to enhance the network's ability to identify and remove rain streaks accurately. Similarly, Ruan et al.~\cite{ruan2024distill} use a spiking network to reconstruct a rain-free background and extract the physical characteristics of rain. 
Ge et al.~\cite{ge2024neuromorphic} use event signals as prior knowledge to improve dynamic information perception and design a deep unfolding optimization algorithm to construct a de-raining network.
The network combines CNNs for spatial feature extraction and Spiking Neural Networks (SNNs) for temporal dynamics.
This hybrid design allows the network to leverage both dense intensity frames and sparse event streams. It helps to improve rain removal while preserving textures.
Spiking Mutual Enhancement (SME) module in the network selectively focuses on relevant spatio-temporal regions and enhances dynamic feature extraction from event streams.

\subsection{Focus Control}  
\label{subsec:focus_control}

The emergence of event cameras, which capture changes in the scene at a high temporal resolution, opens up new possibilities for addressing the challenges of fast and accurate auto-focus in adverse conditions. 
Bao et al.~\cite{bao2023improving} propose an event-based focusing algorithm by leveraging the symmetrical relationship between event polarities. The results demonstrate that precise focus, with less than one depth of focus, is achieved within $0.004$ seconds on a self-built high-speed focusing platform. 
Bao et al.~\cite{Bao_2025_CVPR} propose a one-step event-driven autofocus algorithm, and they reduce the focusing time and focus error significantly as compared to their prior work. They introduce the Event Laplacian Product (ELP) focus detection function. It combines event data with grayscale Laplacian information and formulates the auto focus problem as a detection task. 
The Laplacian of the grayscale image captures image sharpness and edge information. Specifically, it highlights regions of rapid intensity change, which correspond to edges or fine details in the image.
In the context of autofocus, sharper images have stronger Laplacian responses, while blurred images have weaker or smoother Laplacian values.
ELP is essentially a multiplication of the Laplacian of the grayscale image and the event data, followed by a summation.
The sign of the ELP value changes (a "sign mutation") when the system reaches the focus position.
This mutation is used to detect the focus point in real time, and it helps to achieve one-step autofocus without scanning through a focus stack.
The Laplacian provides spatial texture cues that, when combined with temporal event data, allow the system to determine both the focus position and direction of adjustment.

An event camera also helps in generating all-in-focus images using event signal streams during a continuous focal sweep.
Lou et al.~\cite{lou2023all} introduce a method to generate high-quality all-in-focus images from a single shot, addressing the challenges of traditional focal stack methods that require multiple shots. They propose the concept of an event focal stack, which consists of event streams captured during a continuous focal sweep.
The process involves three main steps: first, automatically selecting the optimal timestamps for refocusing based on the high temporal resolution of event streams; second, using these timestamps and corresponding events to reconstruct a series of refocused images, creating an image focal stack; and finally, merging these refocused images with weights predicted from the images and neighboring events to produce a sharp, all-in-focus image. The extension of this approach is proposed by Teng et al.~\cite{teng2024hybrid}, which includes multiple improvements like reformulating the all-in-focus imaging pipeline, handling an arbitrary number of image focal stacks for merging.

\begin{figure*}[!t]
    \centering
    \includegraphics[width=0.60\textwidth]{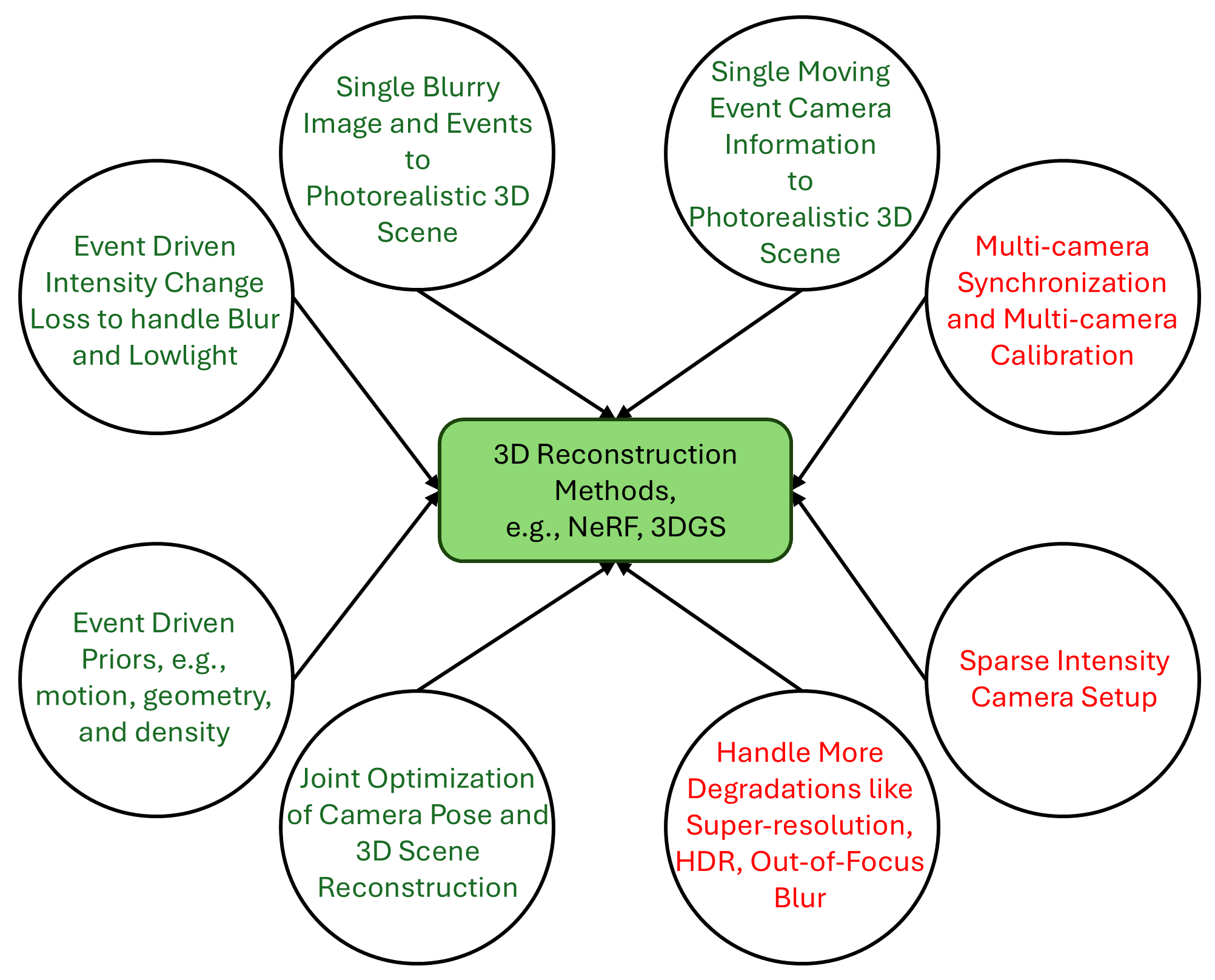}
    \caption{The summary of recent works in photorealistic 3D reconstruction using NeRF and 3DGS with the help of Events. The areas marked as Green show the already explored areas in NeRF and 3DGS. The red-marked areas show the future potential use cases of NeRF and 3DGS.}
    \label{fig:3D}
\end{figure*}

%% file: 50_3dreconstruction.tex
\section{3D Reconstruction} 
\label{sec:3d_recon}

Event cameras register asynchronous per-pixel brightness changes at high rates, which makes them ideal for observing fast motion with minimal motion blur. Due to the high temporal resolution of the asynchronous visual information acquisition, the output of these sensors is ideally suited for dynamic 3D reconstruction like N-ocular 3D reconstruction algorithm for event-based vision data~\cite{carneiro2013event}, 3D reconstruction using data from a stereo event-camera rig in static scenes~\cite{zhou2018semi}, reliable 3D hand mesh reconstruction~\cite{jiang2024complementing}, 3D Hand Sequence Recovery~\cite{park20243d} etc. An event camera has the unique ability to respond to edges of a captured scene, providing geometric information without preprocessing, and continuously measures as the sensor moves. Rebecq et al.~\cite{rebecq2018emvs} exploit this and introduce an event-based multi-view stereo (EMVS) algorithm to estimate semi-dense 3D structures from a known camera trajectory. Baudron et al.~\cite{baudron2020e3d} present an event-to-silhouette (E2S) neural network module that converts event frames into silhouettes and includes neural branches for camera pose regression. 
Traditional photometric stereo techniques require capturing multiple high dynamic range images under different lighting conditions, which limits their speed and real-time applicability. EventPS~\cite{yu2024eventps} overcomes these limitations by estimating surface normals directly from radiance changes detected by the event camera. 

Along with the advancement of event cameras' suitability in different 3D reconstruction domains, photorealistic 3D reconstruction with events gained momentum. Figure~\ref{fig:3D} shows the overall progress of photorealistic 3D reconstruction of events and the novel research opportunities that still persist. Accurate photorealistic 3D reconstruction often requires the fusion of information from multiple viewpoints. 
Real-world deployment of Neural Radiance Fields (NeRF) and 3D Gaussian Splatting (3DGS) methods faces a set of challenges that impact the fidelity and robustness of scene recovery. These include issues such as synchronization and calibration in multi-camera setups, robustness to noise and lighting variations, and limitations in resolution and focus. 
Event cameras, with their unique sensing modality, provide valuable capabilities to address many of these challenges.
As event cameras operate asynchronously and produce per-pixel timestamped events at microsecond resolution, they offer fine-grained temporal information that can be used to achieve highly accurate multi-camera synchronization. This enables consistent spatial alignment and helps prevent artifacts caused by temporal misalignment.
Their unique characteristics enable them to complement conventional frame-based methods, particularly in scenarios with fast motion, varying illumination, and sensor noise. 
Recent works have explored the integration of event data into neural representations for photorealistic 3D reconstruction, including NeRF and 3DGS.

\noindent
\textbf{Advancement of NeRF Models with Events.} 
Rudnev et al.~\cite{rudnev2023eventnerf} developed the first approach in the domain of dense and photorealistic 3D reconstruction using a single color event camera. They used only event camera data for photorealistic 3D reconstruction. 
After that, the advancement in the direction of RGB and event camera information fusion has gained momentum to improve the 3D reconstruction performance of the NeRF model using additional event information. This helps to address multiple challenges of 3D reconstruction, achieving both image deblurring and high-quality sharp NeRF reconstruction from blurry input images, which is common in real-world scenarios due to moving camera and objects~\cite{qi2023e2nerf, klenk2023nerf, rudnev2024dynamic}. 
Qi et al.~\cite{qi2023e2nerf} introduce a blur rendering loss and an event rendering loss to guide the network by modeling the real blur process and event generation process, respectively.
Low et al.~\cite{low2024deblur} propose the Deblur e-NeRF method, which addresses this by incorporating a physically accurate pixel bandwidth model to account for event motion blur and introducing a threshold-normalized total variation loss to improve the regularization of large textureless patches. 
Physically accurate pixel bandwidth refers to a detailed model of how an event camera pixel responds to changes in light intensity over time, based on the actual analog circuitry of the sensor. 
In low-light or high-speed scenarios, limited bandwidth causes motion blur because the pixel cannot react quickly enough, and it leads to delayed or missing events. 
The Deblur e-NeRF paper models this behavior using a cascade of low-pass filters that simulate realistic event simulation and improved NeRF reconstruction.
Li et al.~\cite{li2024benerf} explore the possibility of recovering NeRF from a single blurry image and its corresponding event stream.  
To capture the blurry image, both the RGB camera and the event camera are assumed to follow the same continuous camera motion trajectory during the exposure period.
The proposed method jointly learns the NeRF and the camera's continuous motion trajectory by minimizing the difference between synthesized and real measurements of both RGB camera data and event camera data. 
Qi et al.~\cite{qi2024deblurring} propose EBAD-NeRF, a method designed to enhance NeRF by addressing motion blur issues in low-light and high-speed scenarios, and introduce an intensity-change-metric event loss and a photometric blur loss to model camera motion blur explicitly. 

Other than handling the blurry and low-light scenes, the pose estimation is a challenge in NeRF models. Ma et al.~\cite{ma2023deformable} address challenges such as unknown absolute radiance at event locations and unknown camera poses during events. Their proposed method jointly optimizes the camera poses and the radiance field by leveraging the asynchronous stream of events and calibrating sparse RGB frames. 
Bhattacharya et al.~\cite{bhattacharya2024evdnerf} present an event-based dynamic NeRF (EvDNeRF), which is designed to faithfully reconstruct event streams in scenes with both rigid and non-rigid deformations, which are often too fast to capture with standard cameras. The approach improves test-time predictions of events at fine time resolutions by training on varied batch sizes of events.
Qi et al.~\cite{qi20243} propose a camera pose estimation framework to generalize the method to practical applications. They also introduce two novel losses: an event-enhanced blur rendering loss and an event rendering loss, which model the real blur and event generation processes, respectively.  
Feng et al.~\cite{feng2025ae} introduce a joint pose-NeRF training framework that includes a pose correction module to enhance the robustness of 3D reconstructions from inaccurate camera poses. They address the challenges of reconstructing NeRF from event data captured under non-ideal conditions, such as non-uniform event sequences, noisy poses, and varying scene scales. 
Ev-NeRF~\cite{hwang2023ev} leverages the multi-view consistency of NeRF to provide a self-supervision signal that filters out spurious measurements and extracts consistent underlying structures from the noisy event input. Instead of using posed images like traditional NeRF, Ev-NeRF uses event measurements along with sensor movements to create an integrated neural volume. 

The additional prior derived from the event information helps to achieve a good photorealistic 3D reconstruction.
Cannici et al.~\cite{cannici2024mitigating} combine model-based priors and learning-based modules to enhance NeRF reconstructions using event data and blurry frames. It explicitly models the blur formation process and uses the event double integral as an additional prior. Additionally, the method incorporates an end-to-end learnable response function to adapt to real event-camera sensor non-idealities. 
To overcome the challenges of existing methods that often rely on dense and low-noise event streams, Robust e-NeRF~\cite{low2023robust} incorporates a realistic event generation model that accounts for intrinsic parameters and non-idealities, such as pixel-to-pixel threshold variation. Additionally, it employs a pair of normalized reconstruction losses that generalize effectively to arbitrary speed profiles and intrinsic parameter values without prior knowledge. 
Wang et al.~\cite{wang2024physical} utilize motion, geometry, and density priors to impose strong physical constraints, which significantly improve the robustness and efficiency of 3D scene reconstruction. This method leverages a density-guided patch-based sampling strategy, which accelerates the training process and enhances the expression of local geometries. 

\noindent
\textbf{Advancement of 3DGS Models with Events.}
In contrast to NeRF’s implicit representation, 3D Gaussian Splatting (3DGS) uses explicit 3D Gaussians to model scene geometry and appearance, enabling faster and often more interpretable reconstructions.
Traditional 3DGS algorithms suffer in motion blur scenarios and depend heavily on sharp images and accurate camera poses. It is challenging to obtain in real-world scenarios with different real-world non-ideal settings. However, the incorporation of event information in 3DGS helps to improve the robustness of photorealistic 3D reconstruction and makes 3D reconstruction possible in different challenging scenarios. 
In this direction, Yura et al.~\cite{yura2024eventsplat} use prior knowledge encoded in an event-to-video model for initializing the optimization process of 3DGS and employ spline interpolation to obtain high-quality poses along the event camera trajectory. This method enhances reconstruction quality from fast-moving cameras while overcoming the computational limitations traditionally associated with event-based NeRF methods. 
Huang et al.~\cite{Huang_2025_CVPR} propose IncEventGS, an incremental 3DGS reconstruction algorithm using only a single event camera and without any information about camera pose. IncEventGS divides incoming event streams into chunks and models the camera trajectory as a continuous function. It randomly selects two close timestamps and integrates the corresponding event data. Using 3DGS, it renders two brightness images at the respective poses and minimizes the photometric loss between the synthesized and observed events. During initialization, a pre-trained depth estimation model infers depth from the rendered images to bootstrap the system.
Weng et al.~\cite{weng2024eadeblur} employ an Adaptive Deviation Estimator (ADE) network to estimate Gaussian center deviations and utilize novel loss functions to achieve sharp 3D reconstructions in real time. 
In 3DGS, scenes are represented using Gaussian ellipsoids and each with a center position in 3D space. When input images are blurry due to camera motion, the initial positions of these Gaussians, which are estimated from blurry images, are inaccurate, and this leads to poor 3D reconstructions.
ADE network corrects these inaccuracies and estimates how much each Gaussian center should be deviated using the original Gaussian position and the camera poses estimated from sharp latent images, which are generated using event data.
The novel loss function related to Gaussian center deviation is designed to simulate the motion blur process during exposure and guide the network to learn accurate Gaussian deviations.
This approach demonstrates significant performance improvements over traditional 3DGS methods, particularly in scenarios with high-speed motion and low-light conditions. 
Yu et al.~\cite{yu2024evagaussians} model the formation process of motion-blurred images and guide the deblurring reconstruction of 3DGS by jointly optimizing 3DGS parameters and recovering camera motion trajectories during exposure. 
Event-3DGS~\cite{han2025event} framework processes event data directly and reconstructs 3D scenes by optimizing both scenario and sensor parameters. Key components include a high-pass filter-based photovoltage estimation module to reduce noise and an event-based 3D reconstruction loss to enhance reconstruction quality.
Lee et al.~\cite{Lee_2025_CVPR} propose DiET-GS, which is a diffusion prior and event stream-driven motion deblurring in 3DGS. 
The diffusion prior leverages a pretrained diffusion model to guide the reconstruction toward more natural-looking images, compensating for artifacts introduced by event-based deblurring.
DiET-GS addresses the concerns of inaccurate color and lost fine-grained details that are present in existing solutions. They constrain 3DGS with the event double integral to produce accurate color and details. They also leverage diffusion prior to enhance the fine edge details.

\begin{table*}[h]
\centering
\begin{tabular}{p{0.20\textwidth} | p{0.30\textwidth} | p{0.40\textwidth}}
        \hline
        \textbf{Dataset}  & \textbf{Tasks} & \textbf{Description} \\
        \hline
        LED \cite{duan2024led} 
            & Event denoising, Low light enhancement, video reconstruction
            & $3000$ sequences of high resolution ($1200 \times 680$) paired dataset \\
        EventAID \cite{duan2025eventaid}
            & Image and video enhancement, HDR conversion
            & $1600+$sec of capture with different texture and motion parameters. \\
        RLED \cite{liu2024seeing}
            & Low light enhancement, Frame interpolation, video reconstruction
            & $64200$ aligned image and low-light event pairs\\
        DSEC \cite{gehrig2021dsec}
            & Super-resolution, video reconstruction, Occlusion removal
            & $53$ sequences stereo RGB-Event camera pair along with LIDAR and GPS measurements. \\
        MVSEC \cite{zhu2018multivehicle}
            & HDR conversion, Video reconstruction
            & Stereo capture of gray scale images, event stream, and IMU readings \\
        VECtor \cite{gao2022vector}
            & Occlusion removal, Frame interpolation
            & Stereo capture of RGB event pair, depth sensor, and IMU measurements \\
        IJRR \cite{mueggler2017event}
            & Deblurring, Frame interpolation
            & $24$fps RG images with asynchronous event pair and IMU data \\
        HQF \cite{stoffregen2020reducing}
            & Video reconstruction, Super-resolution
            & Event camera dataset with two independent sensors \\
        TemMat \cite{bao2024temporal}
            & Video reconstruction, Frame interpolation, Focus control
            & Low light high-dynamic range dataset \\
        HighREV @ NITRE2025 \cite{sun2023event}
            & Frame interpolation, Motion deblurring
            & Paired event - RGB data \\
        Erf-X170FPS \cite{kim2023event}
            & Frame interpolation, Video interpolation, Super-resolution
            & High resolution high-fps video frames paired with event data of extremely large motion scenes \\
        CED \cite{scheerlinck2019ced}
            & Colorization, Super resolution, HDR conversion
            & Color event dataset contining $50$min of paired footage \\
        BlinkFlow \cite{li2023blinkflow}
            & Frame interpolation, Video reconstruction, Motion deblurring
            & Simulator for fast generation of event-based optical flow and associated dataset \\
        TUM-VIE \cite{klenk2021tum}
            & 3D reconstruction, Video reconstruction, Super resolution
            & Stereo event camera data from a large variety of handheld and head-mounted sequences in indoor and outdoor environments \\
        EDS \cite{hidalgo2022event}
            & 3D reconstruction, Video, reconstruction, Occlusion removal
            & Paired event-rgb data along with IMU measurements \\
        AE-NeRF \cite{feng2025ae}
            & 3D reconstruction, HDR conversion, Occlusion removal
            & Synthetic dataset based on an improved version of ESIM \\
        PAEv3d \cite{wang2024physical}
            & 3D reconstruction, Occlusion removal, Video reconstruction
            &  Large scale dataset containing $101$ obejcts along with depth maps\\
        EvaGaussians \cite{yu2024evagaussians}
            & Deblurring, Video reconstruction, Frame interpolation 
            & Indoor and outdoor synthetic scenes, and six 3D objects created using Blender, $5$ real world scenes \\
        Real-World-Blur \cite{qi2023e2nerf}
            & Deblurring, 3D reconstruction, Video reconstruction
            & Color event camera dataset of $5$ low-light blur scenes \\
        Real-World-Challenge \cite{qi2023e2nerf}
            & Deblurring, 3D reconstruction, Video reconstruction
            &  Color event camera dataset of $5$ low-light blur scenes \\
        Dynamic EventNeRF \cite{rudnev2023eventnerf}
            & 3D reconstruction, Video reconstruction, Super resolution
            & Multi-view event stream dataset with 6 event cameras \\
        CHMD \cite{liu2025timetracker}
            & Frame interpolation
            &  Aligned dataset of $90$ sequences of high-speed non-linear moving targets \\
        ClerMotion \cite{chen2025repurposing}
            & Frame interpolation
            & Real-world dataset with object and camera motion \\
        \hline
\end{tabular}
\caption{List of openly available datasets with event data stream}
\label{tab:datasets}
\end{table*}

\noindent
\textbf{Other Possibilities in NeRF and 3DGS with Events.}
Apart from already explored non-ideal situations like low-light and motion blur, where event information is useful, it can also be extended to other non-ideal capture scenarios like low-resolution capture, Out-of-focus blur, depth-of-field limitations, multi-camera synchronization, and multi-camera calibration. 
Events, due to their high temporal resolution, can effectively capture fine-grained changes in scene structure. When fused with intensity data or used to guide neural reconstruction models, they enable spatial super-resolution~\cite{han2021evintsr, kai2023video, yu2023learning, lu2023learning, liu2024disentangled, kai2024evtexture, xiao2024event}, improving both texture fidelity and geometric precision.
Out-of-focus blur and depth-of-field limitations are common in real-world intensity captures, particularly when using low-cost or wide-aperture lenses. In such cases, frame-based 3D reconstruction pipelines struggle to resolve scene details due to the loss of high-frequency information. Event cameras, being unaffected by traditional focus constraints, continue to respond to intensity changes regardless of depth-of-field issues~\cite{bao2023improving, lou2023all, teng2024hybrid}. 
Precise calibration of intrinsic and extrinsic parameters is critical for reliable multi-view 3D reconstruction. In conventional systems, calibration relies on detecting specific patterns or correspondences across synchronized frames, which may fail under low-resolution, low-light, fast motion, or in texture-less regions. Event cameras offer dense, temporally continuous observations of scene changes, which can be used to extract motion or edge-based features robustly across different views. This facilitates both offline and online calibration, improving the geometric consistency of reconstructed scenes.
In summary, event-enhanced methods hold significant promise for making photorealistic 3D reconstruction more robust, scalable, and adaptable to real-world conditions.

%% file: 80_datasets.tex
\section{A Survey of Datasets}  
\label{sec:datasets}
In this section, we present a curated list of event-based video datasets, highlighting their importance in advancing tasks such as video reconstruction, frame interpolation, and enhancement under challenging conditions. The table summarizes each dataset by outlining the target tasks and key characteristics, including resolution, sensor modalities, and duration. To provide additional clarity, we briefly categorize datasets based on their primary applications and data composition. 

Despite the growing number of event-based video datasets, several limitations persist that hinder broader applicability and model generalization. Many datasets lack sufficient diversity in scenes, often focusing on constrained environments with limited variation in object types, motion patterns, and backgrounds. Additionally, most datasets provide minimal or no semantic-level annotations, limiting their usefulness for tasks requiring high-level understanding, such as object recognition or scene parsing. There is also a noticeable scarcity of color event datasets, with the majority relying solely on grayscale event streams. Multi-view event datasets, which are crucial for tasks such as 3D reconstruction and view synthesis, remain underrepresented. Furthermore, the range of lighting conditions captured is often narrow, with few datasets systematically covering challenging scenarios such as extreme low light, high dynamic range, or rapid illumination changes. Addressing these gaps is essential for advancing event-based video research and enabling more robust and versatile models.

%% file: 90_conclusion.tex
\section{Conclusion and Research Opportunities}    \label{sec:concluding_remarks}

Event cameras have emerged as a transformative sensing modality, offering high temporal resolution, low latency, and high dynamic range. This survey has comprehensively reviewed their integration with traditional frame-based systems for visual media restoration and 3D reconstruction, highlighting three key domains: temporal enhancement, spatial enhancement, and 3D reconstruction.
In the temporal domain, event cameras enable high-fidelity video reconstruction, frame interpolation, and motion deblurring, especially under fast motion and low-light conditions. Their asynchronous nature allows for precise motion capture and temporal continuity, overcoming limitations of conventional cameras.
In the spatial domain, event data contributes significantly to super-resolution, HDR imaging, low-light enhancement, occlusion removal, and artifact reduction. By capturing fine-grained changes in brightness, event cameras complement RGB data to restore texture and detail in degraded scenes.
In the 3D domain, event cameras facilitate dynamic and photorealistic 3D reconstruction, especially in challenging scenarios involving motion blur, low light, and sparse data. Their integration with NeRF and 3D Gaussian Splatting (3DGS) models has opened new avenues for real-time, high-quality scene modeling.

To further advance the field and guide future research, we propose the following research opportunities:

\begin{enumerate}
\item \textbf{Event-Driven Multi-Modal Fusion:}
Develop robust fusion frameworks that dynamically balance RGB, depth, and event modalities for real-time restoration and reconstruction.

\item \textbf{Low-Resource and Edge Deployment:}
Design lightweight architectures optimized for mobile and embedded platforms for event-based processing in resource-constrained environments.

\item \textbf{Self-Supervised and Unsupervised Learning:}
Explore domain adaptation, contrastive learning, and generative models to reduce reliance on labeled data and improve generalization across scenes to solve the scarcity of event data.

\item \textbf{Event-Based Calibration and Synchronization:}
Investigate calibration-free multi-camera setups using event streams for robust synchronization and alignment in dynamic environments.

\item \textbf{Event-Guided Generative Models:}
Integrate event data with diffusion models for high-quality synthesis and restoration under extreme conditions.

\item \textbf{Event-Based Depth and Focus Estimation:}
Advance depth-from-events and focus control algorithms for applications in robotics, AR/VR, and computational photography.

\item \textbf{Cross-Domain Applications:}
Apply event-based restoration techniques to domains such as medical imaging, remote sensing, and industrial inspection.

\item \textbf{Color Event Cameras:}
Expand research into color event data for improved visual media restoration and reconstruction.

\item \textbf{Benchmarking and Dataset Expansion:}
Create diverse, annotated, and multi-view datasets covering varied lighting, motion, and semantic contexts to support reproducible research.

\item \textbf{Security and Robustness:}
Address vulnerabilities in event-based systems, including adversarial attacks and backdoor threats, to ensure safe deployment in critical applications.

\end{enumerate}

By consolidating recent progress and outlining future directions, this survey aims to serve as a foundational resource for researchers and practitioners seeking to harness the full potential of event cameras in visual media restoration and 3D reconstruction.